%% file: talbot_etal_IROS_2018.tex
\documentclass[letterpaper, 10 pt, conference]{ieeeconf}  

\IEEEoverridecommandlockouts                              

\makeatletter
\let\NAT@parse\undefined                                  
\makeatother

\newcommand{\IEEEwarning}{
  \onecolumn
  \noindent \textcopyright{} 2018 IEEE. Personal use of this material is permitted. Permission from IEEE must be obtained for all other uses, in any current or future media, including reprinting / republishing this material for advertising or promotional purposes, creating new collective works, for resale or redistribution to servers or lists, or reuse of any copyrighted component of this work in other works

  \bigskip \noindent Article submitted to the \textit{International Conference on Intelligent Robots and Systems (IROS) 2018}.
  \twocolumn
  \newpage
}

\usepackage{silence}
\usepackage{xcolor}
\usepackage{soul}
\usepackage[hidelinks]{hyperref}

\usepackage{subcaption}
\usepackage{amsmath}
\usepackage{bm}
\usepackage{algorithm}
\usepackage{algpseudocode}

\usepackage{tikz}
\usepackage{pgfplots}
\usepgfplotslibrary{colormaps,fillbetween}
\pgfplotsset{compat=newest}

\usepackage[noadjust]{cite}

\title{\LARGE \bf OpenSeqSLAM2.0: An Open Source Toolbox for Visual Place Recognition Under Changing Conditions}

\author{
  Ben Talbot$^{1}$, Sourav Garg$^{2}$, and Michael Milford$^{2}$\\
  \thanks{$^{1}$ The authors are with the School of Electrical Engineering and Computer Science, Queensland University of Technology (QUT), Brisbane, Australia b.talbot@qut.edu.au}%
  \thanks{$^{2}$ The authors are with the ARC Centre of Excellence for Robotic Vision, Queensland University of Technology (QUT), Brisbane, Australia. http://roboticvision.org/ email: michael.milford@qut.edu.au}%
}

\begin{document}

\IEEEwarning

\maketitle
\thispagestyle{empty}
\pagestyle{empty}

\input{./source/abstract.tex}
\input{./source/intro.tex}
\input{./source/background.tex}
\input{./source/method.tex}
\input{./source/expdesign.tex}
\input{./source/results.tex}
\input{./source/conclusion.tex}

\bibliographystyle{IEEEtran}
\bibliography{bib/google,bib/custom}
\end{document}

%% file: source/abstract.tex
\begin{abstract}
  Visually recognising a traversed route --- regardless of whether seen during the day or night, in clear or inclement conditions, or in summer or winter --- is an important capability for navigating robots. Since SeqSLAM was introduced in 2012, a large body of work has followed exploring how robotic systems can use the algorithm to meet the challenges posed by navigation in changing environmental conditions. The following paper describes OpenSeqSLAM2.0, a fully open source toolbox for visual place recognition under changing conditions. Beyond the benefits of open access to the source code, OpenSeqSLAM2.0 provides a number of tools to facilitate exploration of the visual place recognition problem and interactive parameter tuning. Using the new open source platform, it is shown for the first time how comprehensive parameter characterisations provide new insights into many of the system components previously presented in ad hoc ways and provide users with a guide to what system component options should be used under what circumstances and why.
\end{abstract}

%% file: source/intro.tex
\section{Introduction}

For navigating robots to become commonplace, strong and robust solutions to the visual place recognition problem --- the ability to recognise a previously visited place --- must be established. While presenting results that demonstrate place recognition for a specific scenario with hand chosen parameters is important, it is also important to develop characterisations of the relationship between parameter values and performance under a wide range of configurations. For place recognition systems to be safely deployed on live robotic systems, parameterisation must be based on a more rigorous understanding than experience-based intuition.

SeqSLAM, a sequence-based method for visual place recognition, has been followed by a wide body of literature since first being introduced by Milford \& Wyeth in 2012 \cite{milford2012seqslam}. The ensuing literature to date has mostly focused on improvements to the algorithm's performance \cite{chen2014convolutional,pepperell2014all,garg2017improving,milford2013vision,neubert2015superpixel,milford2015sequence}, efficiency \cite{siam2017fast,arroyo2015towards,wang2015improved,milford2014condition,chen2015distance}, and a wide range of other contributions \cite{hansen2014visual,milford2015towards,bai2017cnn,sunderhauf2015place,neubert2015local,sunderhauf2015performance,chen2017only}, with minimal focus on how to maximise general performance of the existing algorithm or even the influence exhibited on system performance by each parameter. To maximise the performance of any future visual place recognition system which builds on SeqSLAM, it is crucial that it is first understood how to maximise the performance of the underlying algorithm based on the user's desired deployment scenario.

\begin{figure}[t]
  \centering
  \includegraphics[width=\columnwidth]{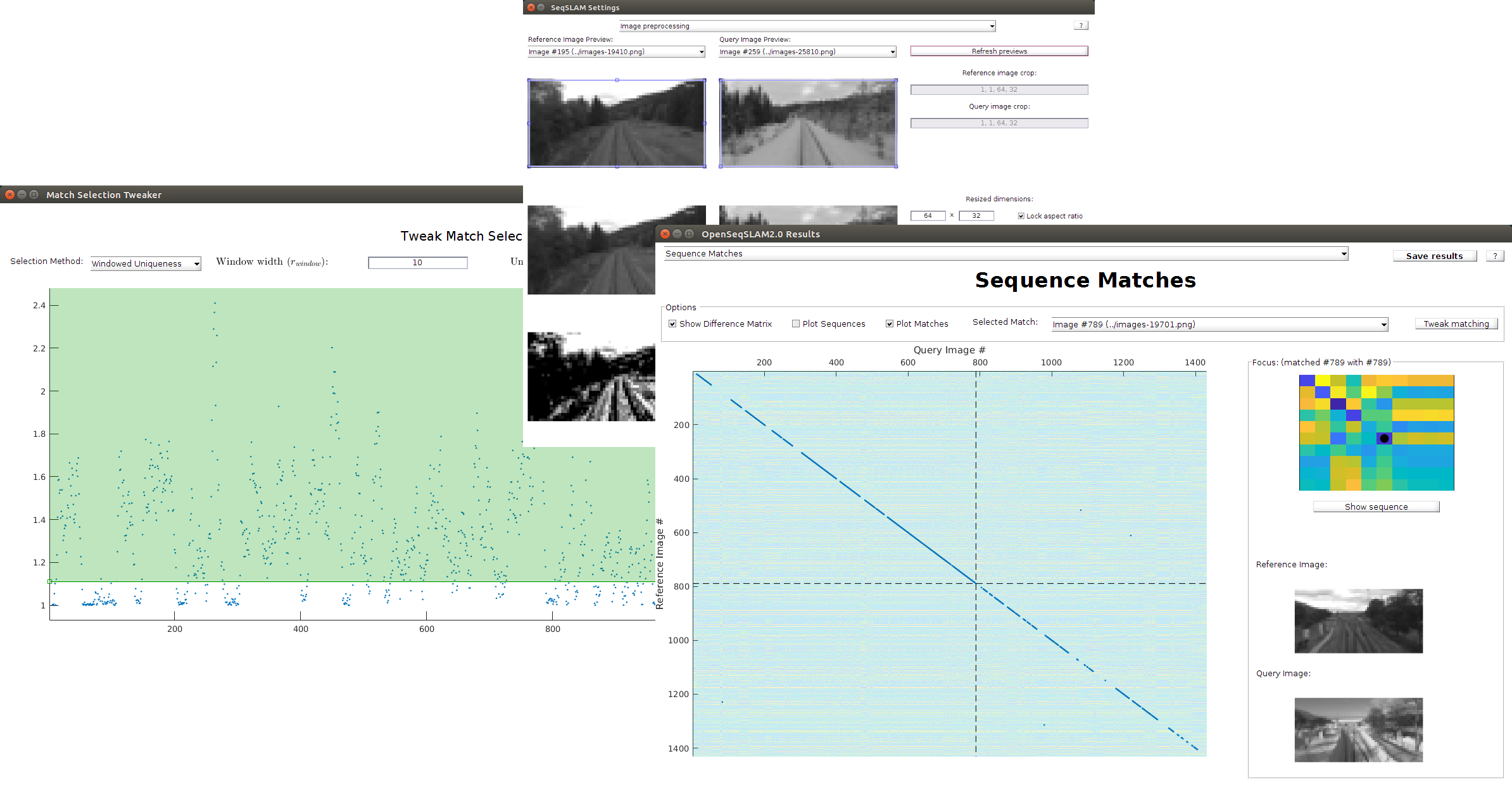}
  \caption{Snapshots from one of the many visual interfaces in the OpenSeqSLAM2.0 toolbox. The snapshots show the ability of the toolbox to dynamically reconfigure threshold parameters post execution, and investigate individual sequence matches from both image and matrix centric perspectives.}
  \label{fig:intro_highlights}
\end{figure}

The following paper presents OpenSeqSLAM2.0, shown in Figure \ref{fig:intro_highlights} and available at \url{http://tiny.cc/openseqslam2}, as a highly configurable open source toolbox for using SeqSLAM. The toolbox facilitates detailed and dynamic exploration of results, as well as characterising the effect of parameters on performance. Using the novel OpenSeqSLAM2.0 software, the paper presents the following contributions:
\begin{itemize}
  \item a comparative summary of a number of different sequence search and match selection techniques that have only previously been presented in isolation;
  \item detailed performance characterisations for numerous variations to the original SeqSLAM algorithm; and
  \item an open source toolbox for the robotics community, boasting the ability to graphically configure parameters, dynamically reconfigure parameters, auto-optimise thresholds to maximise performance, profile performance via precision-recall, and sweep parameter values.
\end{itemize}

The remainder of the paper proceeds as follows. Section \ref{sec:background} provides a brief discussion of visual place recognition approaches, with a focus on performance characterisations and existing open source software. The algorithm implemented in OpenSeqSLAM2.0 is then discussed in Section \ref{sec:method}, including all implemented variations to the approach. In addition, the toolbox is demonstrated with detailed performance characterisation tests which have their experimental design and results outlined in Sections \ref{sec:expdesign} and \ref{sec:results} respectively. To conclude, the major insights provided by the parameter characterisations are presented.

%% file: source/background.tex
\section{Background}
\label{sec:background}

OpenSeqSLAM2.0 approaches the visual place recognition problem using the SeqSLAM method first introduced by Milford et al. \cite{milford2012seqslam}. SeqSLAM is one of many methods available for working with the visual place recognition problem, and all approaches exhibit their own strengths and weaknesses. The ensuing section explores the types and instances of visual place recognition methods, discusses published performance characterisations of existing systems, and lists the existing relevant open source software packages.

\subsection{Visual Place Recognition Approaches}

Visual place recognition has received significant attention in robotics, with a range of approaches presented by the community. Lowry et al. \cite{lowry2016visual} present a discussion of the concepts behind the problem, and detailed survey of the visual place recognition landscape. Instead, the discussion here focuses on only the core points of difference between visual place recognition methods.

Appearance-only based methods like FAB-MAP \cite{cummins2008fab} employ local point features like SIFT \cite{lowe2004distinctive}, ORB \cite{mur2015orb}, etc. in a bag of visual words framework for matching places. However, the features lack robustness towards extreme perceptual changes. Performing visual place recognition under changing environmental conditions typically involves image transformation based on, for example, illumination invariance \cite{mcmanus2014shady}, shadow removal \cite{corke2013dealing}, appearance prediction \cite{neubert2013appearance}, convolutional filtering \cite{sunderhauf2015performance}, etc. Further, sequence-based methods like SeqSLAM \cite{milford2012seqslam}, SMART \cite{pepperell2014all}, and several others \cite{siam2017fast,wang2015improved,vysotska2016lazy,naseer2014robust} sequentially integrate place hypotheses for achieving state-of-the-art performance.

\subsection{Performance Characterisations}

Sequence-based visual place recognition techniques \cite{milford2012seqslam,siam2017fast,wang2015improved,vysotska2016lazy,naseer2014robust}, including SeqSLAM, typically depend on parameter tuning. The parameters are directly related to the different aspects of the challenges faced in recognising places under extreme appearance variations. As a result, the literature has focused on overcoming the challenges posed by appearance variations.

However, the proposed solutions generally deal with only one of the aspects of the problem. Examples include graph-based search strategy \cite{vysotska2016lazy,naseer2014robust} and use of odometry \cite{pepperell2014all} to deal with velocity variability between the compared traverses, deep-learnt features \cite{chen2014convolutional,sunderhauf2015performance} for robust place representation, adapting place neighbourhood range for improved local search \cite{garg2017improving}, etc. The targeted focus in turn limits the characterisation of performance \cite{garg2017improving,neubert2013appearance,skinner2016high,sunderhauf2013we} to merely supporting the research contribution rather than holistically examining parameter influence. The literature does not currently provide exhaustive performance characterisations for most sequence-based techniques, with the performance of SeqSLAM under variations to sequence searching and matching selection poorly understood.

\subsection{Existing Open Source Software}

Over the years the robotics community has released a number of open source software packages for visual place recognition and related technologies. A number of tools exist for visual place recognition including a library for bag of words image conversion \cite{galvez2012bags}, a c++ version of FAB-MAP named openFABMAP \cite{glover2012openfabmap}; an implementation of graph-based matching of image sequences \cite{vysotska2016lazy}; and tools for relocalisation boosted visual place recognition \cite{vysotskarelocalization}. More general tools in relation to visual place recognition include OpenRatSLAM \cite{ball2013openratslam} for the visual appearance based topological SLAM system RatSLAM \cite{milford2004ratslam}; LSD-SLAM for real-time monocular SLAM \cite{engel2014lsd}; and releases of both ORB-SLAM \cite{mur2015orb} and ORB-SLAM2 \cite{mur2017orb}.

For the SeqSLAM method, an earlier open source implentation of OpenSeqSLAM \cite{sunderhauf2013we} was released in 2013 (with other implementations following with various tweaks \cite{siam2017fast}). The first version of OpenSeqSLAM is a thin wrapper around an implementation of SeqSLAM, which is missing a number of features key to analysing performance and optimising parameter selection. Missing features include a graphical interface for interactively configuring parameters, viewing algorithm progress, exploring difference matrices and their enhanced versions, and checking matched sequences; the ability to dynamically reconfigure match selection thresholds post operation; auto-optimisation of thresholds to maximise performance against standard metrics; tools for creating precision recall plots; the option to export matching videos; and an explicit batch operation mode. The OpenSeqSLAM2.0 toolbox provides the highlighted missing features allowing users to rigorously investigate, characterise, and optimise parameter selection.

%% file: source/method.tex
\section{OpenSeqSLAM2.0}
\label{sec:method}

The OpenSeqSLAM2.0 toolbox uses an approach to the visual place recognition problem with the same high level structure as the original SeqSLAM solution \cite{milford2012seqslam}, but adds in a number of configurable variations to steps of the process. Previous work has introduced a number of different methods in isolation, albeit without any analysis of their relative performance. The visual place recognition problem aims to match the images from a query traversal to a reference set of images. For the following section, it is assumed there are $n$ images in the reference traversal and $m$ images for the query.

The SeqSLAM process begins with an image pre-processing stage where all images are transformed into downsampled patch-based representations. Next, a difference matrix is constructed between the two sets of traversal images, and the local contrast within the matrix enhanced. Thirdly, a local sequence search is used to propose match candidates for each of the query dataset images. Finally, match selection is performed to eliminate weak match candidates. Each of the four phases, and the configuration parameters contained in the OpenSeqSLAM2.0 toolbox, are outlined below.

\subsection{Image Pre-processing}

\begin{figure}[t]
  \centering
  \begin{subfigure}[t]{0.48\columnwidth}
    \centering
    \includegraphics[width=\columnwidth]{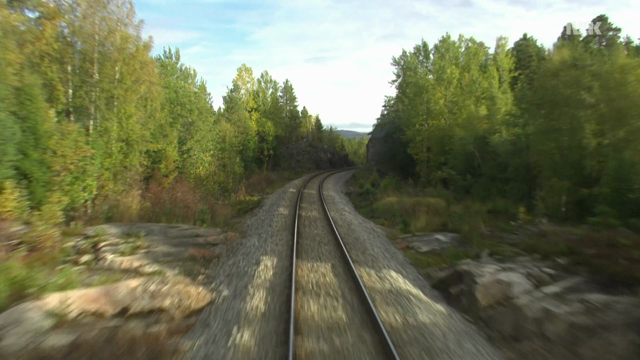}
    \caption{Original Image}
    \label{subfig:prepro_1}
  \end{subfigure}%
  ~
  \begin{subfigure}[t]{0.48\columnwidth}
    \centering
    \includegraphics[width=\columnwidth]{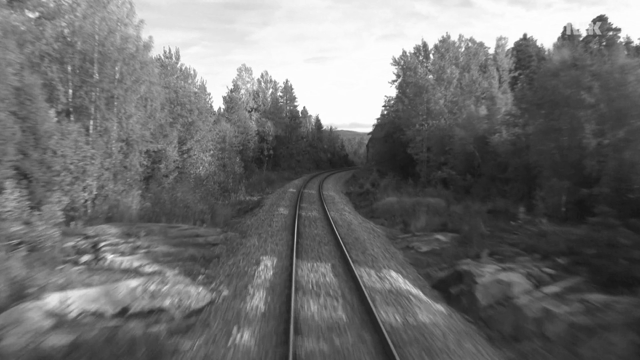}
    \caption{Grayscale Image}
    \label{subfig:prepro_2}
  \end{subfigure}

  \begin{subfigure}[t]{0.48\columnwidth}
    \centering
    \includegraphics[width=\columnwidth]{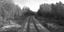}
    \caption{Cropped and Resized Image}
    \label{subfig:prepro_3}
  \end{subfigure}%
  ~
  \begin{subfigure}[t]{0.48\columnwidth}
    \centering
    \includegraphics[width=\columnwidth]{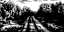}
    \caption{Enhanced Contrast Image}
    \label{subfig:prepro_4}
  \end{subfigure}
  \caption{The ordered pre-processing steps performed on input images by the OpenSeqSLAM2.0 toolbox. All parameters, including crop location, downsized dimensions, aspect ratio, and local contrast enhancement criteria are configurable.}
  \label{fig:method_imprepro}
\end{figure}

The toolbox performs the standard image pre-processing steps, visually outlined in Figure \ref{fig:method_imprepro}. Each image is converted to greyscale, cropped to a region of interest, and downsampled to a resolution consisting of only thousands of pixels. Finally, patch normalisation is performed to enhance the local contrast of the image. The toolbox adds in the ability to adjust all image pre-processing parameters, with visual interfaces to help guide value selection.

\subsection{Difference Matrix Construction}
\label{subsec:normwidth}

Next, a difference matrix of size $n \times m$ is constructed by comparing each patch normalised query image to each patch normalised reference image. The comparison is performed via sum of absolute differences, normalised to the dimensions of the image.

\begin{figure}[t]
  \centering
  \begin{subfigure}[t]{0.5\columnwidth}
    \centering
    \input{./figs/method_base.tex}
    \caption{Before local normalisation}
    \label{subfig:base}
  \end{subfigure}%
  \begin{subfigure}[t]{0.5\columnwidth}
    \centering
    \input{./figs/method_enhance.tex}
    \caption{After local normalisation}
    \label{subfig:enhance}
  \end{subfigure}
  \caption{Local neighbourhood normalisation of the difference vectors for each query image, where the darker colours represent a lower difference score corresponding to a stronger match between reference and query image.}
  \label{fig:method_diffnorm}
\end{figure}
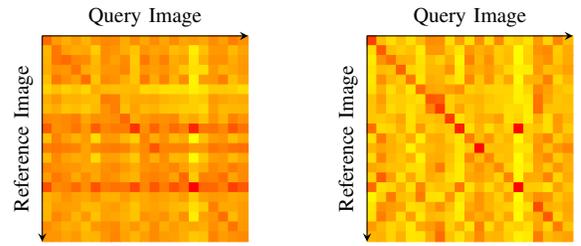

Before the difference matrix can be used to search for place matches, it is first enhanced to facilitate finding local best matches. The enhancement performed --- local neighbourhood normalisation of the difference vector -- is analogous to a 1D version of patch normalisation. For each query image $q$ there is a difference vector $\bm{D^q}$, where each element $r$ corresponds to the difference value for a given reference image. Each element is normalised within a window of size $R_\mathrm{norm}$ to produce a contrast-enhanced image difference vector $\bm{\hat{D}^q}$, where
\begin{equation}
  \bm{\hat{D_r^q}} = \frac{\bm{D_r^q} - \bar{X_r^q}}{\sigma_r^q} \\,
  \label{eq:diffnorm}
\end{equation}
$\bar{X_r^q}$ is the mean within the window, and $\sigma_r^q$ is the standard deviation in the window. The effect of local contrast enhancement on a difference matrix section is shown in Figure \ref{fig:method_diffnorm}. The toolbox allows configuration of the normalisation window, and provides tools to diminish the dominance of outliers in difference matrix visualisation.

\subsection{Local Sequence Recognition}
\label{subsec:search}

With the difference matrix enhanced to emphasise the best matches in local neighbourhoods, the next step in the SeqSLAM process is to search localised sequences for potential matches. Search is performed for every pair of reference image $r$ and query image $q$. In each pair, a search is done for the linear sequence of image pairs that produce the best score under a sequence evaluation metric. The metric is determined by the search method being used.

Previous work has introduced a number of different methods for finding sequences, but without any comparative analysis on what methods might be best. Here for the first time, the OpenSeqSLAM2.0 toolbox presents three different methods of search for local sequence recognition in the same framework. The methods are referred to as trajectory-based \cite{milford2012seqslam}, cone-based \cite{milford2014automated}, and hybrid \cite{chen2014convolutional}. All methods constrain their search to a sequence length $d_s$, and limit the angle of their search for sequences. The angle of search is also often referred to as the velocity through the difference matrix, with limits $v_{\mathrm{min}}$ and $v_{\mathrm{max}}$. Figure \ref{subfig:searchsetup} shows the generic sequence search structure, with the aforementioned parameters highlighted.

\begin{figure}[t]
  \centering
  \begin{subfigure}[t]{0.5\columnwidth}
    \centering
    \input{./figs/method_setup.tex}
    \caption{General search configuration}
    \label{subfig:searchsetup}
  \end{subfigure}%
  ~
  \begin{subfigure}[t]{0.5\columnwidth}
    \centering
    \input{./figs/method_traj.tex}
    \caption{Trajectory-based method}
    \label{subfig:searchtraj}
  \end{subfigure}

  \begin{subfigure}[t]{0.5\columnwidth}
    \centering
    \input{./figs/method_cone.tex}
    \caption{Cone-based method}
    \label{subfig:searchcone}
  \end{subfigure}%
  ~
  \begin{subfigure}[t]{0.5\columnwidth}
    \centering
    \input{./figs/method_hybrid.tex}
    \caption{Hybrid method}
    \label{subfig:searchhybrid}
  \end{subfigure}
  \caption{Visual descriptions for each of the search methods supported by OpenSeqSLAM2.0.}
  \label{fig:method_search}
\end{figure}
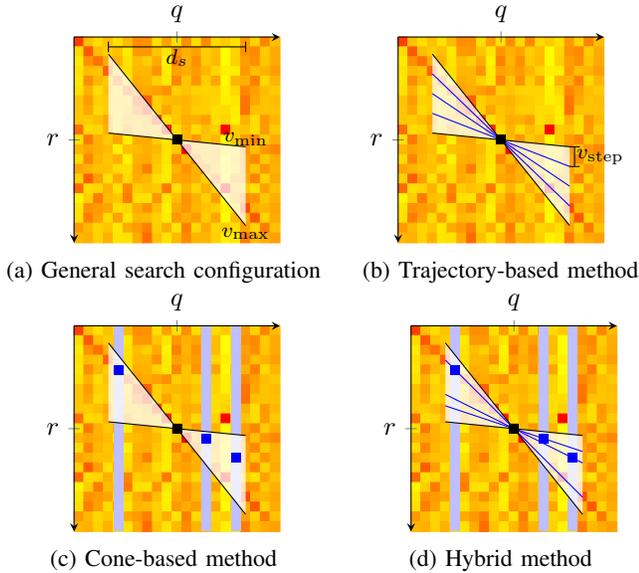

\begin{figure}[t]
  \begin{algorithm}[H]
    {\fontsize{9}{9}\selectfont
      \begin{algorithmic}[1]
        \State $v_\mathrm{steps} \gets \mathrm{linspace}(v_\mathrm{min}, v_\mathrm{max}, v_\mathrm{step})$
        \ForAll{$r,q \gets [1..n], [1..m]$}
        \State $s_\mathrm{best} \gets \infty$
        \For{$v \gets v_\mathrm{steps}$}
        \State $s \gets \mathrm{trajectoryScore}(r, q, v, d_s)$
        \If{$s < s_\mathrm{best}$}
        \State $s_\mathrm{best} \gets s$
        \EndIf
        \EndFor
        \State $s_\mathrm{matrix}(r,q) \gets s_\mathrm{best}$
        \EndFor
      \end{algorithmic}
    }
    \captionof{algorithm}{Trajectory-based search algorithm}
    \label{alg:searchtraj}
  \end{algorithm}
  \vspace{-18pt}
  \begin{algorithm}[H]
    {\fontsize{9}{9}\selectfont
      \begin{algorithmic}[1]
        \ForAll{$r,q \gets [1..n], [1..m]$}
        \State $s \gets 0$
        \ForAll{$c \gets \mathrm{cellsInCones}(r,q)$}
        \If{$\mathrm{isGlobalBest}(c)$}
        \State $s \gets s + 1 / d_s$
        \EndIf
        \EndFor
        \State $s_\mathrm{matrix}(r,q) \gets s$
        \EndFor
      \end{algorithmic}
    }
    \caption{Cone-based search algorithm}
    \label{alg:searchcone}
  \end{algorithm}
  \vspace{-18pt}
  \begin{algorithm}[H]
    {\fontsize{9}{9}\selectfont
      \begin{algorithmic}[1]
        \ForAll{$r,q \gets [1..n], [1..m]$}
        \State $s_\mathrm{best} \gets \infty$
        \ForAll{$c \gets \mathrm{cellsInCones}(r,q)$}
        \If{$\mathrm{isGlobalBest}(c)$}
        \State $v \gets \mathrm{velocity}(r, q, c)$
        \State $s \gets \mathrm{trajectoryScore}(r, q, v, d_s)$
        \If{$s < s_\mathrm{best}$}
        \State $s_\mathrm{best} \gets s$
        \EndIf
        \EndIf
        \EndFor
        \State $s_\mathrm{matrix}(r,q) \gets s_\mathrm{best}$
        \EndFor
      \end{algorithmic}
    }
    \caption{Hybrid search algorithm}
    \label{alg:searchhybrid}
  \end{algorithm}
  \vspace{-18pt}
\end{figure}

Where the methods differ, is in the sequences that they test and how they evaluate their likelihood of representing a match. OpenSeqSLAM2.0 splits the search between looking ahead and behind the query image $q$ (creating the forward and backward facing search cones seen in Figure \ref{subfig:searchsetup}), but other search placements are possible. The explicit details of each search method are shown in Figures \ref{subfig:searchtraj}-\ref{subfig:searchhybrid} and Algorithms \ref{alg:searchtraj}-\ref{alg:searchhybrid} respectively.

Trajectory-based search, used in the first SeqSLAM release, tests sequences of length $d_s$ in steps of $v_\mathrm{step}$ looking for the minimum scoring trajectory for each $r,q$ image pair (where score is the sum of contrast-enhanced difference matrix scores). The method has a fixed number of sequences to search for each image pair, and depending on $v_{\mathrm{step}}$, can be the slowest of the three search methods.

Cone-based search takes a different approach by using the unenhanced difference matrix to search for global best matches in the local neighbourhood. For each $r,q$ image pair, the number of global best matches in the cones of search are counted. The score for each image pair is given as a score between $0$ and $1$, denoting the prevalence of best matches within the search cones ($n_\mathrm{matches}/d_s$, or $3/d_s$ in the case of Figure \ref{subfig:searchcone}).

Lastly, the hybrid search method combines the core principle of the trajectory method with the local search for global best matches of the cone method. Analogous to the first method, a search for global best matches is performed in the local cones, with trajectories drawn through each best match found. The best scoring trajectory is then given as the score for the image pair.

\subsection{Match Selection}
\label{subsec:matches}

As a result of the sequence search process, a matrix of search scores $s_\mathrm{matrix}$ denoting the best score found for each $r,q$ image pairing exists to guide match selection. Each query image $q$ has a vector of scores representing the perceived likelihood that each reference image $r$ could be a match. Match proposals are derived by converting the best score matrix into a vector of scores for each query image's best reference image match. While the entire matches vector could be taken as matches (i.e. every query image is given a match), the approach typically leads to very poor precision results. Instead, a final step is performed to reject poor match proposals by only taking those that pass a given threshold metric. The OpenSeqSLAM2.0 toolbox provides two match selection methods: score thresholding, and windowed uniqueness thresholding. The methods have been published previously, but never had their utility justified.

\subsubsection{Score Thresholding}

The first method is straightforward; all proposed matches with a sequence score below a threshold $\lambda$ are removed from the match list. It is important to note that choosing a threshold can be difficult given that changing sequence length $d_s$ and search method can significantly change the score range. To assist in choosing appropriate thresholds, the toolbox includes an auto-optimisation mode that can choose the threshold which achieves the greatest precision, recall, or $F_1$ score.

\subsubsection{Windowed Uniqueness Thresholding}

\begin{figure}[t]
  \centering
  \input{./figs/method_uniqueness.tex}
  \caption{Calculation of the uniqueness score $\mu_q$ from the reference match scores for image $q$.}
  \label{fig:method_uniqueness}
\end{figure}
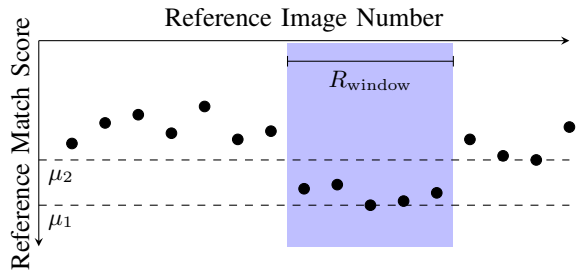

The second mode looks at the uniqueness of a match proposal in its local neighbourhood, rather than the sequence score for the proposal. To get a uniqueness score, the vector of scores for each query image are evaluated. The uniqueness of a match proposal $u_q$ is the percent that the best match's score outperforms the next best match's score, with the next best having to be outside an exclusion window of width $R_\mathrm{window}$ from the best match (i.e. $\mu_1/\mu_2$ in Figure \ref{fig:method_uniqueness}). After a uniqueness score has been given for each query image's match proposal, all proposals that do not have a uniqueness score above $\mu$ are discarded. The uniqueness threshold is significantly easier to configure before running SeqSLAM, with uniqueness scores all typically falling in the same range regardless of parameter values. Auto-optimisation mode is also available in the toolbox for $\mu$ selection.

%% file: figs/method_base.tex
    \begin{tikzpicture}
    \begin{axis}[
    unit vector ratio=1 1 1,
    clip=false,
    colormap/autumn,
    height=\columnwidth,
    width=\columnwidth,
    xtick pos=right,
    axis lines=left,
    ticks=none,
xlabel=\footnotesize Query Image,
ylabel=\footnotesize Reference Image,
xmin=-0.500000, xmax=20.500000, ymin=-0.500000, ymax=20.500000]
\addplot [matrix plot, point meta=explicit] coordinates {
(0,0) [81.386230] (0,1) [130.659668] (0,2) [127.881348] (0,3) [123.510254] (0,4) [135.900879] (0,5) [148.994629] (0,6) [124.295898] (0,7) [132.053711] (0,8) [101.212402] (0,9) [71.244141] (0,10) [137.417480] (0,11) [93.498047] (0,12) [133.416504] (0,13) [97.935547] (0,14) [114.371582] (0,15) [61.677734] (0,16) [127.989258] (0,17) [131.592285] (0,18) [129.577148] (0,19) [102.609375] (0,20) [108.768555] 

(1,0) [111.576172] (1,1) [89.311523] (1,2) [88.416016] (1,3) [106.405273] (1,4) [90.348633] (1,5) [137.010742] (1,6) [111.046387] (1,7) [123.780762] (1,8) [104.515625] (1,9) [105.177246] (1,10) [97.459961] (1,11) [90.074707] (1,12) [101.474609] (1,13) [103.993652] (1,14) [89.629883] (1,15) [101.432129] (1,16) [119.611816] (1,17) [112.892578] (1,18) [100.190918] (1,19) [113.289551] (1,20) [108.859863] 

(2,0) [125.754395] (2,1) [116.104004] (2,2) [81.201660] (2,3) [110.543457] (2,4) [105.883301] (2,5) [147.617676] (2,6) [128.784180] (2,7) [139.414062] (2,8) [101.594238] (2,9) [101.944336] (2,10) [114.657715] (2,11) [94.613281] (2,12) [116.871582] (2,13) [108.691406] (2,14) [100.627441] (2,15) [96.555664] (2,16) [107.194336] (2,17) [115.152832] (2,18) [109.755859] (2,19) [124.979492] (2,20) [119.586914] 

(3,0) [113.940430] (3,1) [123.100586] (3,2) [99.500000] (3,3) [88.048828] (3,4) [117.078125] (3,5) [139.299805] (3,6) [127.747559] (3,7) [134.090332] (3,8) [93.510742] (3,9) [88.335449] (3,10) [126.874023] (3,11) [98.070801] (3,12) [124.534180] (3,13) [110.553223] (3,14) [121.209961] (3,15) [80.855957] (3,16) [116.956543] (3,17) [116.469727] (3,18) [129.464355] (3,19) [120.003418] (3,20) [111.289551] 

(4,0) [128.517090] (4,1) [120.112793] (4,2) [101.892090] (4,3) [122.520996] (4,4) [83.511230] (4,5) [150.207520] (4,6) [131.840820] (4,7) [143.373047] (4,8) [119.620605] (4,9) [110.111328] (4,10) [113.104004] (4,11) [117.708008] (4,12) [117.080566] (4,13) [119.662109] (4,14) [103.267090] (4,15) [103.134766] (4,16) [120.058594] (4,17) [132.821777] (4,18) [110.369141] (4,19) [135.575195] (4,20) [130.310547] 

(5,0) [109.415039] (5,1) [144.056641] (5,2) [132.939453] (5,3) [129.389648] (5,4) [139.266602] (5,5) [124.732422] (5,6) [135.791504] (5,7) [133.402832] (5,8) [105.633789] (5,9) [72.711426] (5,10) [154.462891] (5,11) [100.348145] (5,12) [153.561523] (5,13) [118.439941] (5,14) [120.297852] (5,15) [64.564941] (5,16) [109.498535] (5,17) [130.824219] (5,18) [137.297363] (5,19) [109.743652] (5,20) [128.897949] 

(6,0) [101.603027] (6,1) [103.552246] (6,2) [104.878418] (6,3) [110.058105] (6,4) [102.251465] (6,5) [133.167480] (6,6) [92.484375] (6,7) [106.379883] (6,8) [106.384277] (6,9) [98.554688] (6,10) [112.503418] (6,11) [95.748047] (6,12) [115.816895] (6,13) [105.765625] (6,14) [102.147949] (6,15) [91.936523] (6,16) [125.763672] (6,17) [117.866699] (6,18) [116.792969] (6,19) [105.907227] (6,20) [110.147461] 

(7,0) [102.927246] (7,1) [108.371582] (7,2) [116.135254] (7,3) [109.550293] (7,4) [106.396973] (7,5) [135.869629] (7,6) [95.598633] (7,7) [94.653320] (7,8) [102.743652] (7,9) [91.651367] (7,10) [112.692871] (7,11) [92.520508] (7,12) [114.978027] (7,13) [100.046875] (7,14) [107.068848] (7,15) [85.232422] (7,16) [126.509766] (7,17) [113.981934] (7,18) [119.695312] (7,19) [101.587891] (7,20) [98.870117] 

(8,0) [122.208496] (8,1) [124.186035] (8,2) [102.862793] (8,3) [98.754395] (8,4) [124.912598] (8,5) [133.204590] (8,6) [129.380859] (8,7) [136.031250] (8,8) [79.356934] (8,9) [92.638672] (8,10) [130.710449] (8,11) [96.853516] (8,12) [126.857910] (8,13) [112.550781] (8,14) [121.986816] (8,15) [86.253906] (8,16) [113.978516] (8,17) [124.279785] (8,18) [127.777344] (8,19) [120.930664] (8,20) [116.744141] 

(9,0) [104.125000] (9,1) [139.364258] (9,2) [143.386719] (9,3) [140.816406] (9,4) [140.627930] (9,5) [130.530273] (9,6) [121.911621] (9,7) [117.870605] (9,8) [123.348633] (9,9) [44.210449] (9,10) [145.961914] (9,11) [114.257324] (9,12) [143.936523] (9,13) [123.324707] (9,14) [128.882812] (9,15) [59.977051] (9,16) [127.520996] (9,17) [140.500977] (9,18) [138.775879] (9,19) [110.596191] (9,20) [130.227051] 

(10,0) [120.382812] (10,1) [108.792969] (10,2) [89.941406] (10,3) [108.655273] (10,4) [91.745117] (10,5) [152.627930] (10,6) [120.723145] (10,7) [132.913574] (10,8) [111.846680] (10,9) [105.882324] (10,10) [96.101562] (10,11) [101.795410] (10,12) [104.229492] (10,13) [107.521973] (10,14) [100.037109] (10,15) [98.689941] (10,16) [121.321777] (10,17) [121.476562] (10,18) [108.030762] (10,19) [125.605957] (10,20) [115.899902] 

(11,0) [101.916504] (11,1) [118.353027] (11,2) [111.965332] (11,3) [114.798340] (11,4) [122.854980] (11,5) [150.399902] (11,6) [119.414062] (11,7) [127.280273] (11,8) [103.447754] (11,9) [76.602539] (11,10) [123.889160] (11,11) [67.060547] (11,12) [124.663574] (11,13) [100.713867] (11,14) [95.587402] (11,15) [68.052734] (11,16) [113.500000] (11,17) [116.965332] (11,18) [113.973633] (11,19) [103.037109] (11,20) [103.016602] 

(12,0) [120.397949] (12,1) [113.905762] (12,2) [95.669434] (12,3) [107.081543] (12,4) [96.988770] (12,5) [157.345215] (12,6) [128.892578] (12,7) [138.882812] (12,8) [105.308105] (12,9) [102.723633] (12,10) [106.258301] (12,11) [98.461914] (12,12) [103.949707] (12,13) [107.111328] (12,14) [100.860840] (12,15) [92.217773] (12,16) [117.643555] (12,17) [122.289551] (12,18) [106.918945] (12,19) [124.124023] (12,20) [113.919922] 

(13,0) [102.754395] (13,1) [133.798340] (13,2) [123.339355] (13,3) [121.160645] (13,4) [128.519043] (13,5) [158.425293] (13,6) [133.464844] (13,7) [140.214844] (13,8) [100.858887] (13,9) [81.655273] (13,10) [132.835449] (13,11) [97.181641] (13,12) [131.616699] (13,13) [99.010742] (13,14) [116.254395] (13,15) [65.650391] (13,16) [124.001953] (13,17) [130.480957] (13,18) [127.069336] (13,19) [112.725586] (13,20) [114.831055] 

(14,0) [118.482422] (14,1) [127.166016] (14,2) [116.277344] (14,3) [140.670898] (14,4) [103.548828] (14,5) [157.660156] (14,6) [135.148926] (14,7) [145.921387] (14,8) [130.114258] (14,9) [94.313965] (14,10) [121.967773] (14,11) [97.738770] (14,12) [127.532227] (14,13) [116.094238] (14,14) [80.376953] (14,15) [84.548340] (14,16) [113.536621] (14,17) [134.475586] (14,18) [105.211426] (14,19) [123.493652] (14,20) [125.383301] 

(15,0) [113.777344] (15,1) [166.342773] (15,2) [160.283203] (15,3) [157.348633] (15,4) [170.060547] (15,5) [167.942383] (15,6) [146.272949] (15,7) [145.529785] (15,8) [121.866211] (15,9) [34.814941] (15,10) [178.600586] (15,11) [111.560059] (15,12) [174.630859] (15,13) [133.601074] (15,14) [143.566406] (15,15) [11.937988] (15,16) [142.492676] (15,17) [156.899414] (15,18) [165.814941] (15,19) [119.072754] (15,20) [141.519043] 

(16,0) [121.092773] (16,1) [137.528320] (16,2) [126.309570] (16,3) [155.978516] (16,4) [113.147461] (16,5) [160.099609] (16,6) [142.484863] (16,7) [152.610840] (16,8) [138.046875] (16,9) [90.755371] (16,10) [136.197266] (16,11) [110.567871] (16,12) [141.232422] (16,13) [123.316895] (16,14) [95.445312] (16,15) [78.012207] (16,16) [116.940918] (16,17) [143.127930] (16,18) [117.158691] (16,19) [125.719238] (16,20) [137.068848] 

(17,0) [102.996094] (17,1) [111.029297] (17,2) [106.171875] (17,3) [111.383789] (17,4) [110.741211] (17,5) [142.866211] (17,6) [111.234863] (17,7) [115.791504] (17,8) [108.577148] (17,9) [86.567871] (17,10) [113.479492] (17,11) [88.296387] (17,12) [114.595703] (17,13) [100.784668] (17,14) [98.553711] (17,15) [79.512207] (17,16) [111.019043] (17,17) [94.375000] (17,18) [109.547363] (17,19) [105.362793] (17,20) [96.822754] 

(18,0) [114.784180] (18,1) [109.827148] (18,2) [105.420898] (18,3) [126.961914] (18,4) [93.415039] (18,5) [143.721680] (18,6) [118.982910] (18,7) [132.782715] (18,8) [128.046875] (18,9) [100.097168] (18,10) [108.555664] (18,11) [96.233887] (18,12) [111.330078] (18,13) [110.134277] (18,14) [84.322266] (18,15) [94.279785] (18,16) [113.098145] (18,17) [121.129883] (18,18) [83.721191] (18,19) [118.355957] (18,20) [113.958496] 

(19,0) [93.561035] (19,1) [132.443848] (19,2) [123.768066] (19,3) [119.890137] (19,4) [135.922363] (19,5) [145.356934] (19,6) [124.191406] (19,7) [130.370117] (19,8) [96.279785] (19,9) [68.495117] (19,10) [135.319824] (19,11) [86.685547] (19,12) [133.572754] (19,13) [93.687500] (19,14) [118.244629] (19,15) [54.266602] (19,16) [122.813477] (19,17) [120.232910] (19,18) [133.464844] (19,19) [87.343750] (19,20) [105.416992] 

(20,0) [106.759277] (20,1) [117.720215] (20,2) [112.624512] (20,3) [110.193848] (20,4) [118.518066] (20,5) [153.994629] (20,6) [122.091797] (20,7) [129.271484] (20,8) [108.586426] (20,9) [87.066406] (20,10) [120.679199] (20,11) [83.251953] (20,12) [119.416504] (20,13) [103.075195] (20,14) [101.724121] (20,15) [69.807617] (20,16) [118.160156] (20,17) [112.785645] (20,18) [112.843750] (20,19) [110.384766] (20,20) [101.398438] 

};

\end{axis}
\end{tikzpicture}

%% file: figs/method_enhance.tex
    \begin{tikzpicture}
    \begin{axis}[
    unit vector ratio=1 1 1,
    clip=false,
    colormap/autumn,
    height=\columnwidth,
    width=\columnwidth,
    xtick pos=right,
    axis lines=left,
    ticks=none,
xlabel=\footnotesize Query Image,
ylabel=\footnotesize Reference Image,
xmin=-0.500000, xmax=20.500000, ymin=-0.500000, ymax=20.500000]
\addplot [matrix plot, point meta=explicit] coordinates {
(0,0) [0.998296] (0,1) [1.013897] (0,2) [1.015145] (0,3) [1.014410] (0,4) [1.014964] (0,5) [1.014205] (0,6) [1.011417] (0,7) [1.011526] (0,8) [1.011723] (0,9) [1.003046] (0,10) [1.014108] (0,11) [1.010009] (0,12) [1.013780] (0,13) [1.008474] (0,14) [1.012710] (0,15) [1.002577] (0,16) [1.016219] (0,17) [1.014725] (0,18) [1.014195] (0,19) [1.007308] (0,20) [1.010102] 

(1,0) [1.010999] (1,1) [1.000908] (1,2) [1.006437] (1,3) [1.010528] (1,4) [1.006062] (1,5) [1.009502] (1,6) [1.007622] (1,7) [1.008825] (1,8) [1.012086] (1,9) [1.013684] (1,10) [1.004226] (1,11) [1.008272] (1,12) [1.005382] (1,13) [1.009688] (1,14) [1.001324] (1,15) [1.013879] (1,16) [1.012399] (1,17) [1.006357] (1,18) [1.004510] (1,19) [1.010313] (1,20) [1.009426] 

(2,0) [1.013905] (2,1) [1.010869] (2,2) [1.003306] (2,3) [1.011294] (2,4) [1.010189] (2,5) [1.013841] (2,6) [1.012707] (2,7) [1.013418] (2,8) [1.010883] (2,9) [1.012982] (2,10) [1.009611] (2,11) [1.009471] (2,12) [1.010043] (2,13) [1.011083] (2,14) [1.007801] (2,15) [1.012971] (2,16) [1.004218] (2,17) [1.007595] (2,18) [1.007975] (2,19) [1.013065] (2,20) [1.012758] 

(3,0) [1.011150] (3,1) [1.012264] (3,2) [1.009318] (3,3) [1.001575] (3,4) [1.011980] (3,5) [1.010970] (3,6) [1.012312] (3,7) [1.012197] (3,8) [1.008525] (3,9) [1.009573] (3,10) [1.011976] (3,11) [1.010701] (3,12) [1.011589] (3,13) [1.011165] (3,14) [1.014103] (3,15) [1.009491] (3,16) [1.010665] (3,17) [1.008305] (3,18) [1.013620] (3,19) [1.011606] (3,20) [1.009549] 

(4,0) [1.014166] (4,1) [1.011124] (4,2) [1.009582] (4,3) [1.012820] (4,4) [1.004050] (4,5) [1.014824] (4,6) [1.013157] (4,7) [1.014104] (4,8) [1.014312] (4,9) [1.013950] (4,10) [1.008707] (4,11) [1.015843] (4,12) [1.009415] (4,13) [1.013786] (4,14) [1.008725] (4,15) [1.014057] (4,16) [1.011569] (4,17) [1.013947] (4,18) [1.007763] (4,19) [1.015021] (4,20) [1.014627] 

(5,0) [1.010469] (5,1) [1.015216] (5,2) [1.014485] (5,3) [1.014060] (5,4) [1.014607] (5,5) [1.003063] (5,6) [1.014101] (5,7) [1.012326] (5,8) [1.011212] (5,9) [1.007499] (5,10) [1.015552] (5,11) [1.011188] (5,12) [1.015998] (5,13) [1.014000] (5,14) [1.013465] (5,15) [1.005372] (5,16) [1.004330] (5,17) [1.013319] (5,18) [1.014587] (5,19) [1.008871] (5,20) [1.014429] 

(6,0) [1.005992] (6,1) [1.007168] (6,2) [1.010570] (6,3) [1.010250] (6,4) [1.009143] (6,5) [1.008256] (6,6) [1.000655] (6,7) [1.004778] (6,8) [1.011365] (6,9) [1.012348] (6,10) [1.009210] (6,11) [1.010517] (6,12) [1.009383] (6,13) [1.008763] (6,14) [1.009113] (6,15) [1.012318] (6,16) [1.014063] (6,17) [1.009142] (6,18) [1.010196] (6,19) [1.007226] (6,20) [1.008919] 

(7,0) [1.006410] (7,1) [1.007096] (7,2) [1.012439] (7,3) [1.010103] (7,4) [1.009825] (7,5) [1.008970] (7,6) [1.001539] (7,7) [0.997641] (7,8) [1.010288] (7,9) [1.011264] (7,10) [1.008961] (7,11) [1.009354] (7,12) [1.009065] (7,13) [1.004591] (7,14) [1.010221] (7,15) [1.011112] (7,16) [1.014402] (7,17) [1.006413] (7,18) [1.010839] (7,19) [1.004666] (7,20) [1.002661] 

(8,0) [1.014035] (8,1) [1.011552] (8,2) [1.009015] (8,3) [1.006562] (8,4) [1.012678] (8,5) [1.007821] (8,6) [1.012583] (8,7) [1.012602] (8,8) [0.998037] (8,9) [1.011758] (8,10) [1.012278] (8,11) [1.010540] (8,12) [1.011573] (8,13) [1.012080] (8,14) [1.013852] (8,15) [1.011936] (8,16) [1.006914] (8,17) [1.010865] (8,18) [1.012645] (8,19) [1.012524] (8,20) [1.011191] 

(9,0) [1.007752] (9,1) [1.014632] (9,2) [1.015961] (9,3) [1.015311] (9,4) [1.014909] (9,5) [1.006350] (9,6) [1.010823] (9,7) [1.008276] (9,8) [1.014192] (9,9) [0.995093] (9,10) [1.014685] (9,11) [1.014463] (9,12) [1.014594] (9,13) [1.015198] (9,14) [1.015230] (9,15) [1.003974] (9,16) [1.014625] (9,17) [1.015570] (9,18) [1.015352] (9,19) [1.009127] (9,20) [1.014403] 

(10,0) [1.014198] (10,1) [1.007100] (10,2) [1.004849] (10,3) [1.008097] (10,4) [1.005460] (10,5) [1.012434] (10,6) [1.010296] (10,7) [1.011682] (10,8) [1.011758] (10,9) [1.013981] (10,10) [1.004590] (10,11) [1.012023] (10,12) [1.005802] (10,13) [1.009632] (10,14) [1.008600] (10,15) [1.013939] (10,16) [1.010857] (10,17) [1.008871] (10,18) [1.007212] (10,19) [1.014570] (10,20) [1.010268] 

(11,0) [1.006229] (11,1) [1.009443] (11,2) [1.009961] (11,3) [1.009250] (11,4) [1.011566] (11,5) [1.011293] (11,6) [1.009905] (11,7) [1.010210] (11,8) [1.009121] (11,9) [1.009897] (11,10) [1.010400] (11,11) [0.992653] (11,12) [1.010294] (11,13) [1.006829] (11,14) [1.008095] (11,15) [1.009897] (11,16) [1.006735] (11,17) [1.007086] (11,18) [1.009091] (11,19) [1.004643] (11,20) [1.005745] 

(12,0) [1.013738] (12,1) [1.008132] (12,2) [1.006412] (12,3) [1.007600] (12,4) [1.006772] (12,5) [1.012816] (12,6) [1.011593] (12,7) [1.012283] (12,8) [1.009524] (12,9) [1.013627] (12,10) [1.007046] (12,11) [1.011058] (12,12) [1.005730] (12,13) [1.009587] (12,14) [1.009431] (12,15) [1.013295] (12,16) [1.009653] (12,17) [1.009849] (12,18) [1.007567] (12,19) [1.013608] (12,20) [1.009994] 

(13,0) [1.005511] (13,1) [1.012267] (13,2) [1.011959] (13,3) [1.010084] (13,4) [1.012353] (13,5) [1.013004] (13,6) [1.012507] (13,7) [1.012326] (13,8) [1.007847] (13,9) [1.010735] (13,10) [1.011777] (13,11) [1.010630] (13,12) [1.011531] (13,13) [1.005652] (13,14) [1.012565] (13,15) [1.009564] (13,16) [1.012436] (13,17) [1.011442] (13,18) [1.012182] (13,19) [1.008814] (13,20) [1.009776] 

(14,0) [1.013442] (14,1) [1.010903] (14,2) [1.010535] (14,3) [1.013074] (14,4) [1.008425] (14,5) [1.012736] (14,6) [1.013069] (14,7) [1.013807] (14,8) [1.014065] (14,9) [1.012742] (14,10) [1.010078] (14,11) [1.011020] (14,12) [1.010770] (14,13) [1.012347] (14,14) [1.004922] (14,15) [1.012569] (14,16) [1.007919] (14,17) [1.012264] (14,18) [1.008432] (14,19) [1.013278] (14,20) [1.012542] 

(15,0) [1.011980] (15,1) [1.016948] (15,2) [1.016895] (15,3) [1.015632] (15,4) [1.016808] (15,5) [1.015826] (15,6) [1.015584] (15,7) [1.013745] (15,8) [1.012733] (15,9) [0.992399] (15,10) [1.017028] (15,11) [1.014698] (15,12) [1.016955] (15,13) [1.016401] (15,14) [1.016749] (15,15) [0.992130] (15,16) [1.017366] (15,17) [1.016042] (15,18) [1.016882] (15,19) [1.012227] (15,20) [1.015770] 

(16,0) [1.014228] (16,1) [1.012952] (16,2) [1.012536] (16,3) [1.015469] (16,4) [1.009875] (16,5) [1.013208] (16,6) [1.014727] (16,7) [1.015277] (16,8) [1.015590] (16,9) [1.012200] (16,10) [1.012216] (16,11) [1.014428] (16,12) [1.012992] (16,13) [1.014429] (16,14) [1.009003] (16,15) [1.011989] (16,16) [1.009821] (16,17) [1.014018] (16,18) [1.010666] (16,19) [1.013964] (16,20) [1.015037] 

(17,0) [1.007324] (17,1) [1.006513] (17,2) [1.007860] (17,3) [1.007301] (17,4) [1.009460] (17,5) [1.005462] (17,6) [1.003583] (17,7) [1.003227] (17,8) [1.009474] (17,9) [1.011404] (17,10) [1.007749] (17,11) [1.006867] (17,12) [1.007822] (17,13) [1.008207] (17,14) [1.009225] (17,15) [1.012208] (17,16) [1.006124] (17,17) [0.999405] (17,18) [1.008999] (17,19) [1.008169] (17,20) [1.004331] 

(18,0) [1.012891] (18,1) [1.005584] (18,2) [1.006221] (18,3) [1.010774] (18,4) [1.004566] (18,5) [1.007950] (18,6) [1.008646] (18,7) [1.011063] (18,8) [1.013851] (18,9) [1.013883] (18,10) [1.005373] (18,11) [1.010755] (18,12) [1.005779] (18,13) [1.011511] (18,14) [1.005001] (18,15) [1.014355] (18,16) [1.006888] (18,17) [1.009512] (18,18) [1.000923] (18,19) [1.012783] (18,20) [1.010450] 

(19,0) [1.005217] (19,1) [1.012212] (19,2) [1.011625] (19,3) [1.009905] (19,4) [1.013467] (19,5) [1.010881] (19,6) [1.011392] (19,7) [1.011418] (19,8) [1.007130] (19,9) [1.007540] (19,10) [1.012028] (19,11) [1.007083] (19,12) [1.011862] (19,13) [1.006924] (19,14) [1.012863] (19,15) [1.008012] (19,16) [1.011713] (19,17) [1.010130] (19,18) [1.012853] (19,19) [1.004173] (19,20) [1.009048] 

(20,0) [1.010634] (20,1) [1.009021] (20,2) [1.008849] (20,3) [1.007830] (20,4) [1.010001] (20,5) [1.011954] (20,6) [1.010703] (20,7) [1.011142] (20,8) [1.010304] (20,9) [1.011983] (20,10) [1.009120] (20,11) [1.005477] (20,12) [1.009088] (20,13) [1.009628] (20,14) [1.008801] (20,15) [1.011332] (20,16) [1.009406] (20,17) [1.008406] (20,18) [1.008950] (20,19) [1.011361] (20,20) [1.007979] 

};

\end{axis}
\end{tikzpicture}

%% file: figs/method_setup.tex
    \begin{tikzpicture}
    \begin{axis}[
    unit vector ratio=1 1 1,
    clip=false,
    colormap/autumn,
    height=\columnwidth,
    width=\columnwidth,
    xtick pos=right,
    axis lines=left,
    xtick={10.000000},xticklabels={$q$},ytick={10.000000},yticklabels={$r$},xmin=-0.500000, xmax=20.500000, ymin=-0.500000, ymax=20.500000]
\addplot [matrix plot, point meta=explicit] coordinates {
(0,0) [0.998296] (0,1) [1.013897] (0,2) [1.015145] (0,3) [1.014410] (0,4) [1.014964] (0,5) [1.014205] (0,6) [1.011417] (0,7) [1.011526] (0,8) [1.011723] (0,9) [1.003046] (0,10) [1.014108] (0,11) [1.010009] (0,12) [1.013780] (0,13) [1.008474] (0,14) [1.012710] (0,15) [1.002577] (0,16) [1.016219] (0,17) [1.014725] (0,18) [1.014195] (0,19) [1.007308] (0,20) [1.010102] 

(1,0) [1.010999] (1,1) [1.000908] (1,2) [1.006437] (1,3) [1.010528] (1,4) [1.006062] (1,5) [1.009502] (1,6) [1.007622] (1,7) [1.008825] (1,8) [1.012086] (1,9) [1.013684] (1,10) [1.004226] (1,11) [1.008272] (1,12) [1.005382] (1,13) [1.009688] (1,14) [1.001324] (1,15) [1.013879] (1,16) [1.012399] (1,17) [1.006357] (1,18) [1.004510] (1,19) [1.010313] (1,20) [1.009426] 

(2,0) [1.013905] (2,1) [1.010869] (2,2) [1.003306] (2,3) [1.011294] (2,4) [1.010189] (2,5) [1.013841] (2,6) [1.012707] (2,7) [1.013418] (2,8) [1.010883] (2,9) [1.012982] (2,10) [1.009611] (2,11) [1.009471] (2,12) [1.010043] (2,13) [1.011083] (2,14) [1.007801] (2,15) [1.012971] (2,16) [1.004218] (2,17) [1.007595] (2,18) [1.007975] (2,19) [1.013065] (2,20) [1.012758] 

(3,0) [1.011150] (3,1) [1.012264] (3,2) [1.009318] (3,3) [1.001575] (3,4) [1.011980] (3,5) [1.010970] (3,6) [1.012312] (3,7) [1.012197] (3,8) [1.008525] (3,9) [1.009573] (3,10) [1.011976] (3,11) [1.010701] (3,12) [1.011589] (3,13) [1.011165] (3,14) [1.014103] (3,15) [1.009491] (3,16) [1.010665] (3,17) [1.008305] (3,18) [1.013620] (3,19) [1.011606] (3,20) [1.009549] 

(4,0) [1.014166] (4,1) [1.011124] (4,2) [1.009582] (4,3) [1.012820] (4,4) [1.004050] (4,5) [1.014824] (4,6) [1.013157] (4,7) [1.014104] (4,8) [1.014312] (4,9) [1.013950] (4,10) [1.008707] (4,11) [1.015843] (4,12) [1.009415] (4,13) [1.013786] (4,14) [1.008725] (4,15) [1.014057] (4,16) [1.011569] (4,17) [1.013947] (4,18) [1.007763] (4,19) [1.015021] (4,20) [1.014627] 

(5,0) [1.010469] (5,1) [1.015216] (5,2) [1.014485] (5,3) [1.014060] (5,4) [1.014607] (5,5) [1.003063] (5,6) [1.014101] (5,7) [1.012326] (5,8) [1.011212] (5,9) [1.007499] (5,10) [1.015552] (5,11) [1.011188] (5,12) [1.015998] (5,13) [1.014000] (5,14) [1.013465] (5,15) [1.005372] (5,16) [1.004330] (5,17) [1.013319] (5,18) [1.014587] (5,19) [1.008871] (5,20) [1.014429] 

(6,0) [1.005992] (6,1) [1.007168] (6,2) [1.010570] (6,3) [1.010250] (6,4) [1.009143] (6,5) [1.008256] (6,6) [1.000655] (6,7) [1.004778] (6,8) [1.011365] (6,9) [1.012348] (6,10) [1.009210] (6,11) [1.010517] (6,12) [1.009383] (6,13) [1.008763] (6,14) [1.009113] (6,15) [1.012318] (6,16) [1.014063] (6,17) [1.009142] (6,18) [1.010196] (6,19) [1.007226] (6,20) [1.008919] 

(7,0) [1.006410] (7,1) [1.007096] (7,2) [1.012439] (7,3) [1.010103] (7,4) [1.009825] (7,5) [1.008970] (7,6) [1.001539] (7,7) [0.997641] (7,8) [1.010288] (7,9) [1.011264] (7,10) [1.008961] (7,11) [1.009354] (7,12) [1.009065] (7,13) [1.004591] (7,14) [1.010221] (7,15) [1.011112] (7,16) [1.014402] (7,17) [1.006413] (7,18) [1.010839] (7,19) [1.004666] (7,20) [1.002661] 

(8,0) [1.014035] (8,1) [1.011552] (8,2) [1.009015] (8,3) [1.006562] (8,4) [1.012678] (8,5) [1.007821] (8,6) [1.012583] (8,7) [1.012602] (8,8) [0.998037] (8,9) [1.011758] (8,10) [1.012278] (8,11) [1.010540] (8,12) [1.011573] (8,13) [1.012080] (8,14) [1.013852] (8,15) [1.011936] (8,16) [1.006914] (8,17) [1.010865] (8,18) [1.012645] (8,19) [1.012524] (8,20) [1.011191] 

(9,0) [1.007752] (9,1) [1.014632] (9,2) [1.015961] (9,3) [1.015311] (9,4) [1.014909] (9,5) [1.006350] (9,6) [1.010823] (9,7) [1.008276] (9,8) [1.014192] (9,9) [0.995093] (9,10) [1.014685] (9,11) [1.014463] (9,12) [1.014594] (9,13) [1.015198] (9,14) [1.015230] (9,15) [1.003974] (9,16) [1.014625] (9,17) [1.015570] (9,18) [1.015352] (9,19) [1.009127] (9,20) [1.014403] 

(10,0) [1.014198] (10,1) [1.007100] (10,2) [1.004849] (10,3) [1.008097] (10,4) [1.005460] (10,5) [1.012434] (10,6) [1.010296] (10,7) [1.011682] (10,8) [1.011758] (10,9) [1.013981] (10,10) [1.004590] (10,11) [1.012023] (10,12) [1.005802] (10,13) [1.009632] (10,14) [1.008600] (10,15) [1.013939] (10,16) [1.010857] (10,17) [1.008871] (10,18) [1.007212] (10,19) [1.014570] (10,20) [1.010268] 

(11,0) [1.006229] (11,1) [1.009443] (11,2) [1.009961] (11,3) [1.009250] (11,4) [1.011566] (11,5) [1.011293] (11,6) [1.009905] (11,7) [1.010210] (11,8) [1.009121] (11,9) [1.009897] (11,10) [1.010400] (11,11) [0.992653] (11,12) [1.010294] (11,13) [1.006829] (11,14) [1.008095] (11,15) [1.009897] (11,16) [1.006735] (11,17) [1.007086] (11,18) [1.009091] (11,19) [1.004643] (11,20) [1.005745] 

(12,0) [1.013738] (12,1) [1.008132] (12,2) [1.006412] (12,3) [1.007600] (12,4) [1.006772] (12,5) [1.012816] (12,6) [1.011593] (12,7) [1.012283] (12,8) [1.009524] (12,9) [1.013627] (12,10) [1.007046] (12,11) [1.011058] (12,12) [1.005730] (12,13) [1.009587] (12,14) [1.009431] (12,15) [1.013295] (12,16) [1.009653] (12,17) [1.009849] (12,18) [1.007567] (12,19) [1.013608] (12,20) [1.009994] 

(13,0) [1.005511] (13,1) [1.012267] (13,2) [1.011959] (13,3) [1.010084] (13,4) [1.012353] (13,5) [1.013004] (13,6) [1.012507] (13,7) [1.012326] (13,8) [1.007847] (13,9) [1.010735] (13,10) [1.011777] (13,11) [1.010630] (13,12) [1.011531] (13,13) [1.005652] (13,14) [1.012565] (13,15) [1.009564] (13,16) [1.012436] (13,17) [1.011442] (13,18) [1.012182] (13,19) [1.008814] (13,20) [1.009776] 

(14,0) [1.013442] (14,1) [1.010903] (14,2) [1.010535] (14,3) [1.013074] (14,4) [1.008425] (14,5) [1.012736] (14,6) [1.013069] (14,7) [1.013807] (14,8) [1.014065] (14,9) [1.012742] (14,10) [1.010078] (14,11) [1.011020] (14,12) [1.010770] (14,13) [1.012347] (14,14) [1.004922] (14,15) [1.012569] (14,16) [1.007919] (14,17) [1.012264] (14,18) [1.008432] (14,19) [1.013278] (14,20) [1.012542] 

(15,0) [1.011980] (15,1) [1.016948] (15,2) [1.016895] (15,3) [1.015632] (15,4) [1.016808] (15,5) [1.015826] (15,6) [1.015584] (15,7) [1.013745] (15,8) [1.012733] (15,9) [0.992399] (15,10) [1.017028] (15,11) [1.014698] (15,12) [1.016955] (15,13) [1.016401] (15,14) [1.016749] (15,15) [0.992130] (15,16) [1.017366] (15,17) [1.016042] (15,18) [1.016882] (15,19) [1.012227] (15,20) [1.015770] 

(16,0) [1.014228] (16,1) [1.012952] (16,2) [1.012536] (16,3) [1.015469] (16,4) [1.009875] (16,5) [1.013208] (16,6) [1.014727] (16,7) [1.015277] (16,8) [1.015590] (16,9) [1.012200] (16,10) [1.012216] (16,11) [1.014428] (16,12) [1.012992] (16,13) [1.014429] (16,14) [1.009003] (16,15) [1.011989] (16,16) [1.009821] (16,17) [1.014018] (16,18) [1.010666] (16,19) [1.013964] (16,20) [1.015037] 

(17,0) [1.007324] (17,1) [1.006513] (17,2) [1.007860] (17,3) [1.007301] (17,4) [1.009460] (17,5) [1.005462] (17,6) [1.003583] (17,7) [1.003227] (17,8) [1.009474] (17,9) [1.011404] (17,10) [1.007749] (17,11) [1.006867] (17,12) [1.007822] (17,13) [1.008207] (17,14) [1.009225] (17,15) [1.012208] (17,16) [1.006124] (17,17) [0.999405] (17,18) [1.008999] (17,19) [1.008169] (17,20) [1.004331] 

(18,0) [1.012891] (18,1) [1.005584] (18,2) [1.006221] (18,3) [1.010774] (18,4) [1.004566] (18,5) [1.007950] (18,6) [1.008646] (18,7) [1.011063] (18,8) [1.013851] (18,9) [1.013883] (18,10) [1.005373] (18,11) [1.010755] (18,12) [1.005779] (18,13) [1.011511] (18,14) [1.005001] (18,15) [1.014355] (18,16) [1.006888] (18,17) [1.009512] (18,18) [1.000923] (18,19) [1.012783] (18,20) [1.010450] 

(19,0) [1.005217] (19,1) [1.012212] (19,2) [1.011625] (19,3) [1.009905] (19,4) [1.013467] (19,5) [1.010881] (19,6) [1.011392] (19,7) [1.011418] (19,8) [1.007130] (19,9) [1.007540] (19,10) [1.012028] (19,11) [1.007083] (19,12) [1.011862] (19,13) [1.006924] (19,14) [1.012863] (19,15) [1.008012] (19,16) [1.011713] (19,17) [1.010130] (19,18) [1.012853] (19,19) [1.004173] (19,20) [1.009048] 

(20,0) [1.010634] (20,1) [1.009021] (20,2) [1.008849] (20,3) [1.007830] (20,4) [1.010001] (20,5) [1.011954] (20,6) [1.010703] (20,7) [1.011142] (20,8) [1.010304] (20,9) [1.011983] (20,10) [1.009120] (20,11) [1.005477] (20,12) [1.009088] (20,13) [1.009628] (20,14) [1.008801] (20,15) [1.011332] (20,16) [1.009406] (20,17) [1.008406] (20,18) [1.008950] (20,19) [1.011361] (20,20) [1.007979] 

};
\addplot[
    draw=none,
    fill=white,
    fill opacity=0.75
    ] coordinates {
    (3.000000,9.300000) (17.000000,10.700000) (17.000000,18.750000) (3.000000,1.250000)
};
\addplot[name path=A] coordinates {(3.000000,9.300000) (17.000000,10.700000)};
\addplot[name path=B] coordinates {(3.000000,1.250000) (17.000000,18.750000)};
\addplot[draw=none,fill=black] coordinates {(9.500000,9.500000) (10.500000,9.500000) (10.500000,10.500000) (9.500000,10.500000)};
\node [inner sep=0.5] at (axis cs:17.000000,10.700000) [anchor=south] {\footnotesize $v_{\mathrm{min}}$};
\node [inner sep=0.5] at (axis cs:17.000000,18.750000) [anchor=north] {\footnotesize $v_{\mathrm{max}}$};
\addplot[mark=|] coordinates {(3.000000,0.500000) (17.000000,0.500000)};
\node [inner sep=0.5] at (axis cs:10.000000,0.500000) [anchor=north] {\footnotesize $d_s$};

\end{axis}
\end{tikzpicture}

%% file: figs/method_traj.tex
    \begin{tikzpicture}
    \begin{axis}[
    unit vector ratio=1 1 1,
    clip=false,
    colormap/autumn,
    height=\columnwidth,
    width=\columnwidth,
    xtick pos=right,
    axis lines=left,
    xtick={10.000000},xticklabels={$q$},ytick={10.000000},yticklabels={$r$},xmin=-0.500000, xmax=20.500000, ymin=-0.500000, ymax=20.500000]
\addplot [matrix plot, point meta=explicit] coordinates {
(0,0) [0.998296] (0,1) [1.013897] (0,2) [1.015145] (0,3) [1.014410] (0,4) [1.014964] (0,5) [1.014205] (0,6) [1.011417] (0,7) [1.011526] (0,8) [1.011723] (0,9) [1.003046] (0,10) [1.014108] (0,11) [1.010009] (0,12) [1.013780] (0,13) [1.008474] (0,14) [1.012710] (0,15) [1.002577] (0,16) [1.016219] (0,17) [1.014725] (0,18) [1.014195] (0,19) [1.007308] (0,20) [1.010102] 

(1,0) [1.010999] (1,1) [1.000908] (1,2) [1.006437] (1,3) [1.010528] (1,4) [1.006062] (1,5) [1.009502] (1,6) [1.007622] (1,7) [1.008825] (1,8) [1.012086] (1,9) [1.013684] (1,10) [1.004226] (1,11) [1.008272] (1,12) [1.005382] (1,13) [1.009688] (1,14) [1.001324] (1,15) [1.013879] (1,16) [1.012399] (1,17) [1.006357] (1,18) [1.004510] (1,19) [1.010313] (1,20) [1.009426] 

(2,0) [1.013905] (2,1) [1.010869] (2,2) [1.003306] (2,3) [1.011294] (2,4) [1.010189] (2,5) [1.013841] (2,6) [1.012707] (2,7) [1.013418] (2,8) [1.010883] (2,9) [1.012982] (2,10) [1.009611] (2,11) [1.009471] (2,12) [1.010043] (2,13) [1.011083] (2,14) [1.007801] (2,15) [1.012971] (2,16) [1.004218] (2,17) [1.007595] (2,18) [1.007975] (2,19) [1.013065] (2,20) [1.012758] 

(3,0) [1.011150] (3,1) [1.012264] (3,2) [1.009318] (3,3) [1.001575] (3,4) [1.011980] (3,5) [1.010970] (3,6) [1.012312] (3,7) [1.012197] (3,8) [1.008525] (3,9) [1.009573] (3,10) [1.011976] (3,11) [1.010701] (3,12) [1.011589] (3,13) [1.011165] (3,14) [1.014103] (3,15) [1.009491] (3,16) [1.010665] (3,17) [1.008305] (3,18) [1.013620] (3,19) [1.011606] (3,20) [1.009549] 

(4,0) [1.014166] (4,1) [1.011124] (4,2) [1.009582] (4,3) [1.012820] (4,4) [1.004050] (4,5) [1.014824] (4,6) [1.013157] (4,7) [1.014104] (4,8) [1.014312] (4,9) [1.013950] (4,10) [1.008707] (4,11) [1.015843] (4,12) [1.009415] (4,13) [1.013786] (4,14) [1.008725] (4,15) [1.014057] (4,16) [1.011569] (4,17) [1.013947] (4,18) [1.007763] (4,19) [1.015021] (4,20) [1.014627] 

(5,0) [1.010469] (5,1) [1.015216] (5,2) [1.014485] (5,3) [1.014060] (5,4) [1.014607] (5,5) [1.003063] (5,6) [1.014101] (5,7) [1.012326] (5,8) [1.011212] (5,9) [1.007499] (5,10) [1.015552] (5,11) [1.011188] (5,12) [1.015998] (5,13) [1.014000] (5,14) [1.013465] (5,15) [1.005372] (5,16) [1.004330] (5,17) [1.013319] (5,18) [1.014587] (5,19) [1.008871] (5,20) [1.014429] 

(6,0) [1.005992] (6,1) [1.007168] (6,2) [1.010570] (6,3) [1.010250] (6,4) [1.009143] (6,5) [1.008256] (6,6) [1.000655] (6,7) [1.004778] (6,8) [1.011365] (6,9) [1.012348] (6,10) [1.009210] (6,11) [1.010517] (6,12) [1.009383] (6,13) [1.008763] (6,14) [1.009113] (6,15) [1.012318] (6,16) [1.014063] (6,17) [1.009142] (6,18) [1.010196] (6,19) [1.007226] (6,20) [1.008919] 

(7,0) [1.006410] (7,1) [1.007096] (7,2) [1.012439] (7,3) [1.010103] (7,4) [1.009825] (7,5) [1.008970] (7,6) [1.001539] (7,7) [0.997641] (7,8) [1.010288] (7,9) [1.011264] (7,10) [1.008961] (7,11) [1.009354] (7,12) [1.009065] (7,13) [1.004591] (7,14) [1.010221] (7,15) [1.011112] (7,16) [1.014402] (7,17) [1.006413] (7,18) [1.010839] (7,19) [1.004666] (7,20) [1.002661] 

(8,0) [1.014035] (8,1) [1.011552] (8,2) [1.009015] (8,3) [1.006562] (8,4) [1.012678] (8,5) [1.007821] (8,6) [1.012583] (8,7) [1.012602] (8,8) [0.998037] (8,9) [1.011758] (8,10) [1.012278] (8,11) [1.010540] (8,12) [1.011573] (8,13) [1.012080] (8,14) [1.013852] (8,15) [1.011936] (8,16) [1.006914] (8,17) [1.010865] (8,18) [1.012645] (8,19) [1.012524] (8,20) [1.011191] 

(9,0) [1.007752] (9,1) [1.014632] (9,2) [1.015961] (9,3) [1.015311] (9,4) [1.014909] (9,5) [1.006350] (9,6) [1.010823] (9,7) [1.008276] (9,8) [1.014192] (9,9) [0.995093] (9,10) [1.014685] (9,11) [1.014463] (9,12) [1.014594] (9,13) [1.015198] (9,14) [1.015230] (9,15) [1.003974] (9,16) [1.014625] (9,17) [1.015570] (9,18) [1.015352] (9,19) [1.009127] (9,20) [1.014403] 

(10,0) [1.014198] (10,1) [1.007100] (10,2) [1.004849] (10,3) [1.008097] (10,4) [1.005460] (10,5) [1.012434] (10,6) [1.010296] (10,7) [1.011682] (10,8) [1.011758] (10,9) [1.013981] (10,10) [1.004590] (10,11) [1.012023] (10,12) [1.005802] (10,13) [1.009632] (10,14) [1.008600] (10,15) [1.013939] (10,16) [1.010857] (10,17) [1.008871] (10,18) [1.007212] (10,19) [1.014570] (10,20) [1.010268] 

(11,0) [1.006229] (11,1) [1.009443] (11,2) [1.009961] (11,3) [1.009250] (11,4) [1.011566] (11,5) [1.011293] (11,6) [1.009905] (11,7) [1.010210] (11,8) [1.009121] (11,9) [1.009897] (11,10) [1.010400] (11,11) [0.992653] (11,12) [1.010294] (11,13) [1.006829] (11,14) [1.008095] (11,15) [1.009897] (11,16) [1.006735] (11,17) [1.007086] (11,18) [1.009091] (11,19) [1.004643] (11,20) [1.005745] 

(12,0) [1.013738] (12,1) [1.008132] (12,2) [1.006412] (12,3) [1.007600] (12,4) [1.006772] (12,5) [1.012816] (12,6) [1.011593] (12,7) [1.012283] (12,8) [1.009524] (12,9) [1.013627] (12,10) [1.007046] (12,11) [1.011058] (12,12) [1.005730] (12,13) [1.009587] (12,14) [1.009431] (12,15) [1.013295] (12,16) [1.009653] (12,17) [1.009849] (12,18) [1.007567] (12,19) [1.013608] (12,20) [1.009994] 

(13,0) [1.005511] (13,1) [1.012267] (13,2) [1.011959] (13,3) [1.010084] (13,4) [1.012353] (13,5) [1.013004] (13,6) [1.012507] (13,7) [1.012326] (13,8) [1.007847] (13,9) [1.010735] (13,10) [1.011777] (13,11) [1.010630] (13,12) [1.011531] (13,13) [1.005652] (13,14) [1.012565] (13,15) [1.009564] (13,16) [1.012436] (13,17) [1.011442] (13,18) [1.012182] (13,19) [1.008814] (13,20) [1.009776] 

(14,0) [1.013442] (14,1) [1.010903] (14,2) [1.010535] (14,3) [1.013074] (14,4) [1.008425] (14,5) [1.012736] (14,6) [1.013069] (14,7) [1.013807] (14,8) [1.014065] (14,9) [1.012742] (14,10) [1.010078] (14,11) [1.011020] (14,12) [1.010770] (14,13) [1.012347] (14,14) [1.004922] (14,15) [1.012569] (14,16) [1.007919] (14,17) [1.012264] (14,18) [1.008432] (14,19) [1.013278] (14,20) [1.012542] 

(15,0) [1.011980] (15,1) [1.016948] (15,2) [1.016895] (15,3) [1.015632] (15,4) [1.016808] (15,5) [1.015826] (15,6) [1.015584] (15,7) [1.013745] (15,8) [1.012733] (15,9) [0.992399] (15,10) [1.017028] (15,11) [1.014698] (15,12) [1.016955] (15,13) [1.016401] (15,14) [1.016749] (15,15) [0.992130] (15,16) [1.017366] (15,17) [1.016042] (15,18) [1.016882] (15,19) [1.012227] (15,20) [1.015770] 

(16,0) [1.014228] (16,1) [1.012952] (16,2) [1.012536] (16,3) [1.015469] (16,4) [1.009875] (16,5) [1.013208] (16,6) [1.014727] (16,7) [1.015277] (16,8) [1.015590] (16,9) [1.012200] (16,10) [1.012216] (16,11) [1.014428] (16,12) [1.012992] (16,13) [1.014429] (16,14) [1.009003] (16,15) [1.011989] (16,16) [1.009821] (16,17) [1.014018] (16,18) [1.010666] (16,19) [1.013964] (16,20) [1.015037] 

(17,0) [1.007324] (17,1) [1.006513] (17,2) [1.007860] (17,3) [1.007301] (17,4) [1.009460] (17,5) [1.005462] (17,6) [1.003583] (17,7) [1.003227] (17,8) [1.009474] (17,9) [1.011404] (17,10) [1.007749] (17,11) [1.006867] (17,12) [1.007822] (17,13) [1.008207] (17,14) [1.009225] (17,15) [1.012208] (17,16) [1.006124] (17,17) [0.999405] (17,18) [1.008999] (17,19) [1.008169] (17,20) [1.004331] 

(18,0) [1.012891] (18,1) [1.005584] (18,2) [1.006221] (18,3) [1.010774] (18,4) [1.004566] (18,5) [1.007950] (18,6) [1.008646] (18,7) [1.011063] (18,8) [1.013851] (18,9) [1.013883] (18,10) [1.005373] (18,11) [1.010755] (18,12) [1.005779] (18,13) [1.011511] (18,14) [1.005001] (18,15) [1.014355] (18,16) [1.006888] (18,17) [1.009512] (18,18) [1.000923] (18,19) [1.012783] (18,20) [1.010450] 

(19,0) [1.005217] (19,1) [1.012212] (19,2) [1.011625] (19,3) [1.009905] (19,4) [1.013467] (19,5) [1.010881] (19,6) [1.011392] (19,7) [1.011418] (19,8) [1.007130] (19,9) [1.007540] (19,10) [1.012028] (19,11) [1.007083] (19,12) [1.011862] (19,13) [1.006924] (19,14) [1.012863] (19,15) [1.008012] (19,16) [1.011713] (19,17) [1.010130] (19,18) [1.012853] (19,19) [1.004173] (19,20) [1.009048] 

(20,0) [1.010634] (20,1) [1.009021] (20,2) [1.008849] (20,3) [1.007830] (20,4) [1.010001] (20,5) [1.011954] (20,6) [1.010703] (20,7) [1.011142] (20,8) [1.010304] (20,9) [1.011983] (20,10) [1.009120] (20,11) [1.005477] (20,12) [1.009088] (20,13) [1.009628] (20,14) [1.008801] (20,15) [1.011332] (20,16) [1.009406] (20,17) [1.008406] (20,18) [1.008950] (20,19) [1.011361] (20,20) [1.007979] 

};
\addplot[
    draw=none,
    fill=white,
    fill opacity=0.75
    ] coordinates {
    (3.000000,9.300000) (17.000000,10.700000) (17.000000,18.750000) (3.000000,1.250000)
};
\addplot[name path=A] coordinates {(3.000000,9.300000) (17.000000,10.700000)};
\addplot[name path=B] coordinates {(3.000000,1.250000) (17.000000,18.750000)};
\addplot[draw=none,fill=black] coordinates {(9.500000,9.500000) (10.500000,9.500000) (10.500000,10.500000) (9.500000,10.500000)};
\addplot[draw=blue] coordinates {(3.000000,7.287500) (17.000000,12.712500)};
\addplot[draw=blue] coordinates {(3.000000,5.275000) (17.000000,14.725000)};
\addplot[draw=blue] coordinates {(3.000000,3.262500) (17.000000,16.737500)};
\addplot[mark=-] coordinates {(17.500000,10.700000) (17.500000,12.712500)};
\node [inner sep=2.5] at (axis cs:17.000000,11.706250)[anchor=west] {\footnotesize $v_{\mathrm{step}}$};
\addplot[draw=none,fill=black] coordinates {(9.500000,9.500000) (10.500000,9.500000) (10.500000,10.500000) (9.500000,10.500000)};

\end{axis}
\end{tikzpicture}

%% file: figs/method_cone.tex
    \begin{tikzpicture}
    \begin{axis}[
    unit vector ratio=1 1 1,
    clip=false,
    colormap/autumn,
    height=\columnwidth,
    width=\columnwidth,
    xtick pos=right,
    axis lines=left,
    xtick={10.000000},xticklabels={$q$},ytick={10.000000},yticklabels={$r$},xmin=-0.500000, xmax=20.500000, ymin=-0.500000, ymax=20.500000]
\addplot [matrix plot, point meta=explicit] coordinates {
(0,0) [0.998296] (0,1) [1.013897] (0,2) [1.015145] (0,3) [1.014410] (0,4) [1.014964] (0,5) [1.014205] (0,6) [1.011417] (0,7) [1.011526] (0,8) [1.011723] (0,9) [1.003046] (0,10) [1.014108] (0,11) [1.010009] (0,12) [1.013780] (0,13) [1.008474] (0,14) [1.012710] (0,15) [1.002577] (0,16) [1.016219] (0,17) [1.014725] (0,18) [1.014195] (0,19) [1.007308] (0,20) [1.010102] 

(1,0) [1.010999] (1,1) [1.000908] (1,2) [1.006437] (1,3) [1.010528] (1,4) [1.006062] (1,5) [1.009502] (1,6) [1.007622] (1,7) [1.008825] (1,8) [1.012086] (1,9) [1.013684] (1,10) [1.004226] (1,11) [1.008272] (1,12) [1.005382] (1,13) [1.009688] (1,14) [1.001324] (1,15) [1.013879] (1,16) [1.012399] (1,17) [1.006357] (1,18) [1.004510] (1,19) [1.010313] (1,20) [1.009426] 

(2,0) [1.013905] (2,1) [1.010869] (2,2) [1.003306] (2,3) [1.011294] (2,4) [1.010189] (2,5) [1.013841] (2,6) [1.012707] (2,7) [1.013418] (2,8) [1.010883] (2,9) [1.012982] (2,10) [1.009611] (2,11) [1.009471] (2,12) [1.010043] (2,13) [1.011083] (2,14) [1.007801] (2,15) [1.012971] (2,16) [1.004218] (2,17) [1.007595] (2,18) [1.007975] (2,19) [1.013065] (2,20) [1.012758] 

(3,0) [1.011150] (3,1) [1.012264] (3,2) [1.009318] (3,3) [1.001575] (3,4) [1.011980] (3,5) [1.010970] (3,6) [1.012312] (3,7) [1.012197] (3,8) [1.008525] (3,9) [1.009573] (3,10) [1.011976] (3,11) [1.010701] (3,12) [1.011589] (3,13) [1.011165] (3,14) [1.014103] (3,15) [1.009491] (3,16) [1.010665] (3,17) [1.008305] (3,18) [1.013620] (3,19) [1.011606] (3,20) [1.009549] 

(4,0) [1.014166] (4,1) [1.011124] (4,2) [1.009582] (4,3) [1.012820] (4,4) [1.004050] (4,5) [1.014824] (4,6) [1.013157] (4,7) [1.014104] (4,8) [1.014312] (4,9) [1.013950] (4,10) [1.008707] (4,11) [1.015843] (4,12) [1.009415] (4,13) [1.013786] (4,14) [1.008725] (4,15) [1.014057] (4,16) [1.011569] (4,17) [1.013947] (4,18) [1.007763] (4,19) [1.015021] (4,20) [1.014627] 

(5,0) [1.010469] (5,1) [1.015216] (5,2) [1.014485] (5,3) [1.014060] (5,4) [1.014607] (5,5) [1.003063] (5,6) [1.014101] (5,7) [1.012326] (5,8) [1.011212] (5,9) [1.007499] (5,10) [1.015552] (5,11) [1.011188] (5,12) [1.015998] (5,13) [1.014000] (5,14) [1.013465] (5,15) [1.005372] (5,16) [1.004330] (5,17) [1.013319] (5,18) [1.014587] (5,19) [1.008871] (5,20) [1.014429] 

(6,0) [1.005992] (6,1) [1.007168] (6,2) [1.010570] (6,3) [1.010250] (6,4) [1.009143] (6,5) [1.008256] (6,6) [1.000655] (6,7) [1.004778] (6,8) [1.011365] (6,9) [1.012348] (6,10) [1.009210] (6,11) [1.010517] (6,12) [1.009383] (6,13) [1.008763] (6,14) [1.009113] (6,15) [1.012318] (6,16) [1.014063] (6,17) [1.009142] (6,18) [1.010196] (6,19) [1.007226] (6,20) [1.008919] 

(7,0) [1.006410] (7,1) [1.007096] (7,2) [1.012439] (7,3) [1.010103] (7,4) [1.009825] (7,5) [1.008970] (7,6) [1.001539] (7,7) [0.997641] (7,8) [1.010288] (7,9) [1.011264] (7,10) [1.008961] (7,11) [1.009354] (7,12) [1.009065] (7,13) [1.004591] (7,14) [1.010221] (7,15) [1.011112] (7,16) [1.014402] (7,17) [1.006413] (7,18) [1.010839] (7,19) [1.004666] (7,20) [1.002661] 

(8,0) [1.014035] (8,1) [1.011552] (8,2) [1.009015] (8,3) [1.006562] (8,4) [1.012678] (8,5) [1.007821] (8,6) [1.012583] (8,7) [1.012602] (8,8) [0.998037] (8,9) [1.011758] (8,10) [1.012278] (8,11) [1.010540] (8,12) [1.011573] (8,13) [1.012080] (8,14) [1.013852] (8,15) [1.011936] (8,16) [1.006914] (8,17) [1.010865] (8,18) [1.012645] (8,19) [1.012524] (8,20) [1.011191] 

(9,0) [1.007752] (9,1) [1.014632] (9,2) [1.015961] (9,3) [1.015311] (9,4) [1.014909] (9,5) [1.006350] (9,6) [1.010823] (9,7) [1.008276] (9,8) [1.014192] (9,9) [0.995093] (9,10) [1.014685] (9,11) [1.014463] (9,12) [1.014594] (9,13) [1.015198] (9,14) [1.015230] (9,15) [1.003974] (9,16) [1.014625] (9,17) [1.015570] (9,18) [1.015352] (9,19) [1.009127] (9,20) [1.014403] 

(10,0) [1.014198] (10,1) [1.007100] (10,2) [1.004849] (10,3) [1.008097] (10,4) [1.005460] (10,5) [1.012434] (10,6) [1.010296] (10,7) [1.011682] (10,8) [1.011758] (10,9) [1.013981] (10,10) [1.004590] (10,11) [1.012023] (10,12) [1.005802] (10,13) [1.009632] (10,14) [1.008600] (10,15) [1.013939] (10,16) [1.010857] (10,17) [1.008871] (10,18) [1.007212] (10,19) [1.014570] (10,20) [1.010268] 

(11,0) [1.006229] (11,1) [1.009443] (11,2) [1.009961] (11,3) [1.009250] (11,4) [1.011566] (11,5) [1.011293] (11,6) [1.009905] (11,7) [1.010210] (11,8) [1.009121] (11,9) [1.009897] (11,10) [1.010400] (11,11) [0.992653] (11,12) [1.010294] (11,13) [1.006829] (11,14) [1.008095] (11,15) [1.009897] (11,16) [1.006735] (11,17) [1.007086] (11,18) [1.009091] (11,19) [1.004643] (11,20) [1.005745] 

(12,0) [1.013738] (12,1) [1.008132] (12,2) [1.006412] (12,3) [1.007600] (12,4) [1.006772] (12,5) [1.012816] (12,6) [1.011593] (12,7) [1.012283] (12,8) [1.009524] (12,9) [1.013627] (12,10) [1.007046] (12,11) [1.011058] (12,12) [1.005730] (12,13) [1.009587] (12,14) [1.009431] (12,15) [1.013295] (12,16) [1.009653] (12,17) [1.009849] (12,18) [1.007567] (12,19) [1.013608] (12,20) [1.009994] 

(13,0) [1.005511] (13,1) [1.012267] (13,2) [1.011959] (13,3) [1.010084] (13,4) [1.012353] (13,5) [1.013004] (13,6) [1.012507] (13,7) [1.012326] (13,8) [1.007847] (13,9) [1.010735] (13,10) [1.011777] (13,11) [1.010630] (13,12) [1.011531] (13,13) [1.005652] (13,14) [1.012565] (13,15) [1.009564] (13,16) [1.012436] (13,17) [1.011442] (13,18) [1.012182] (13,19) [1.008814] (13,20) [1.009776] 

(14,0) [1.013442] (14,1) [1.010903] (14,2) [1.010535] (14,3) [1.013074] (14,4) [1.008425] (14,5) [1.012736] (14,6) [1.013069] (14,7) [1.013807] (14,8) [1.014065] (14,9) [1.012742] (14,10) [1.010078] (14,11) [1.011020] (14,12) [1.010770] (14,13) [1.012347] (14,14) [1.004922] (14,15) [1.012569] (14,16) [1.007919] (14,17) [1.012264] (14,18) [1.008432] (14,19) [1.013278] (14,20) [1.012542] 

(15,0) [1.011980] (15,1) [1.016948] (15,2) [1.016895] (15,3) [1.015632] (15,4) [1.016808] (15,5) [1.015826] (15,6) [1.015584] (15,7) [1.013745] (15,8) [1.012733] (15,9) [0.992399] (15,10) [1.017028] (15,11) [1.014698] (15,12) [1.016955] (15,13) [1.016401] (15,14) [1.016749] (15,15) [0.992130] (15,16) [1.017366] (15,17) [1.016042] (15,18) [1.016882] (15,19) [1.012227] (15,20) [1.015770] 

(16,0) [1.014228] (16,1) [1.012952] (16,2) [1.012536] (16,3) [1.015469] (16,4) [1.009875] (16,5) [1.013208] (16,6) [1.014727] (16,7) [1.015277] (16,8) [1.015590] (16,9) [1.012200] (16,10) [1.012216] (16,11) [1.014428] (16,12) [1.012992] (16,13) [1.014429] (16,14) [1.009003] (16,15) [1.011989] (16,16) [1.009821] (16,17) [1.014018] (16,18) [1.010666] (16,19) [1.013964] (16,20) [1.015037] 

(17,0) [1.007324] (17,1) [1.006513] (17,2) [1.007860] (17,3) [1.007301] (17,4) [1.009460] (17,5) [1.005462] (17,6) [1.003583] (17,7) [1.003227] (17,8) [1.009474] (17,9) [1.011404] (17,10) [1.007749] (17,11) [1.006867] (17,12) [1.007822] (17,13) [1.008207] (17,14) [1.009225] (17,15) [1.012208] (17,16) [1.006124] (17,17) [0.999405] (17,18) [1.008999] (17,19) [1.008169] (17,20) [1.004331] 

(18,0) [1.012891] (18,1) [1.005584] (18,2) [1.006221] (18,3) [1.010774] (18,4) [1.004566] (18,5) [1.007950] (18,6) [1.008646] (18,7) [1.011063] (18,8) [1.013851] (18,9) [1.013883] (18,10) [1.005373] (18,11) [1.010755] (18,12) [1.005779] (18,13) [1.011511] (18,14) [1.005001] (18,15) [1.014355] (18,16) [1.006888] (18,17) [1.009512] (18,18) [1.000923] (18,19) [1.012783] (18,20) [1.010450] 

(19,0) [1.005217] (19,1) [1.012212] (19,2) [1.011625] (19,3) [1.009905] (19,4) [1.013467] (19,5) [1.010881] (19,6) [1.011392] (19,7) [1.011418] (19,8) [1.007130] (19,9) [1.007540] (19,10) [1.012028] (19,11) [1.007083] (19,12) [1.011862] (19,13) [1.006924] (19,14) [1.012863] (19,15) [1.008012] (19,16) [1.011713] (19,17) [1.010130] (19,18) [1.012853] (19,19) [1.004173] (19,20) [1.009048] 

(20,0) [1.010634] (20,1) [1.009021] (20,2) [1.008849] (20,3) [1.007830] (20,4) [1.010001] (20,5) [1.011954] (20,6) [1.010703] (20,7) [1.011142] (20,8) [1.010304] (20,9) [1.011983] (20,10) [1.009120] (20,11) [1.005477] (20,12) [1.009088] (20,13) [1.009628] (20,14) [1.008801] (20,15) [1.011332] (20,16) [1.009406] (20,17) [1.008406] (20,18) [1.008950] (20,19) [1.011361] (20,20) [1.007979] 

};
\draw [draw=none,fill=blue!25] (3.5,-0.5) rectangle (4.5,20.500000);
\draw [draw=none,fill=blue!25] (12.5,-0.5) rectangle (13.5,20.500000);
\draw [draw=none,fill=blue!25] (15.5,-0.5) rectangle (16.5,20.500000);
\addplot[
    draw=none,
    fill=white,
    fill opacity=0.75
    ] coordinates {
    (3.000000,9.300000) (17.000000,10.700000) (17.000000,18.750000) (3.000000,1.250000)
};
\addplot[name path=A] coordinates {(3.000000,9.300000) (17.000000,10.700000)};
\addplot[name path=B] coordinates {(3.000000,1.250000) (17.000000,18.750000)};
\addplot[draw=none,fill=black] coordinates {(9.500000,9.500000) (10.500000,9.500000) (10.500000,10.500000) (9.500000,10.500000)};
\addplot[draw=none,fill=blue] coordinates {(3.500000,3.500000) (4.500000,3.500000) (4.500000,4.500000) (3.500000,4.500000)};
\addplot[draw=none,fill=blue] coordinates {(12.500000,10.500000) (13.500000,10.500000) (13.500000,11.500000) (12.500000,11.500000)};
\addplot[draw=none,fill=blue] coordinates {(15.500000,12.500000) (16.500000,12.500000) (16.500000,13.500000) (15.500000,13.500000)};

\end{axis}
\end{tikzpicture}

%% file: figs/method_hybrid.tex
    \begin{tikzpicture}
    \begin{axis}[
    unit vector ratio=1 1 1,
    clip=false,
    colormap/autumn,
    height=\columnwidth,
    width=\columnwidth,
    xtick pos=right,
    axis lines=left,
    xtick={10.000000},xticklabels={$q$},ytick={10.000000},yticklabels={$r$},xmin=-0.500000, xmax=20.500000, ymin=-0.500000, ymax=20.500000]
\addplot [matrix plot, point meta=explicit] coordinates {
(0,0) [0.998296] (0,1) [1.013897] (0,2) [1.015145] (0,3) [1.014410] (0,4) [1.014964] (0,5) [1.014205] (0,6) [1.011417] (0,7) [1.011526] (0,8) [1.011723] (0,9) [1.003046] (0,10) [1.014108] (0,11) [1.010009] (0,12) [1.013780] (0,13) [1.008474] (0,14) [1.012710] (0,15) [1.002577] (0,16) [1.016219] (0,17) [1.014725] (0,18) [1.014195] (0,19) [1.007308] (0,20) [1.010102] 

(1,0) [1.010999] (1,1) [1.000908] (1,2) [1.006437] (1,3) [1.010528] (1,4) [1.006062] (1,5) [1.009502] (1,6) [1.007622] (1,7) [1.008825] (1,8) [1.012086] (1,9) [1.013684] (1,10) [1.004226] (1,11) [1.008272] (1,12) [1.005382] (1,13) [1.009688] (1,14) [1.001324] (1,15) [1.013879] (1,16) [1.012399] (1,17) [1.006357] (1,18) [1.004510] (1,19) [1.010313] (1,20) [1.009426] 

(2,0) [1.013905] (2,1) [1.010869] (2,2) [1.003306] (2,3) [1.011294] (2,4) [1.010189] (2,5) [1.013841] (2,6) [1.012707] (2,7) [1.013418] (2,8) [1.010883] (2,9) [1.012982] (2,10) [1.009611] (2,11) [1.009471] (2,12) [1.010043] (2,13) [1.011083] (2,14) [1.007801] (2,15) [1.012971] (2,16) [1.004218] (2,17) [1.007595] (2,18) [1.007975] (2,19) [1.013065] (2,20) [1.012758] 

(3,0) [1.011150] (3,1) [1.012264] (3,2) [1.009318] (3,3) [1.001575] (3,4) [1.011980] (3,5) [1.010970] (3,6) [1.012312] (3,7) [1.012197] (3,8) [1.008525] (3,9) [1.009573] (3,10) [1.011976] (3,11) [1.010701] (3,12) [1.011589] (3,13) [1.011165] (3,14) [1.014103] (3,15) [1.009491] (3,16) [1.010665] (3,17) [1.008305] (3,18) [1.013620] (3,19) [1.011606] (3,20) [1.009549] 

(4,0) [1.014166] (4,1) [1.011124] (4,2) [1.009582] (4,3) [1.012820] (4,4) [1.004050] (4,5) [1.014824] (4,6) [1.013157] (4,7) [1.014104] (4,8) [1.014312] (4,9) [1.013950] (4,10) [1.008707] (4,11) [1.015843] (4,12) [1.009415] (4,13) [1.013786] (4,14) [1.008725] (4,15) [1.014057] (4,16) [1.011569] (4,17) [1.013947] (4,18) [1.007763] (4,19) [1.015021] (4,20) [1.014627] 

(5,0) [1.010469] (5,1) [1.015216] (5,2) [1.014485] (5,3) [1.014060] (5,4) [1.014607] (5,5) [1.003063] (5,6) [1.014101] (5,7) [1.012326] (5,8) [1.011212] (5,9) [1.007499] (5,10) [1.015552] (5,11) [1.011188] (5,12) [1.015998] (5,13) [1.014000] (5,14) [1.013465] (5,15) [1.005372] (5,16) [1.004330] (5,17) [1.013319] (5,18) [1.014587] (5,19) [1.008871] (5,20) [1.014429] 

(6,0) [1.005992] (6,1) [1.007168] (6,2) [1.010570] (6,3) [1.010250] (6,4) [1.009143] (6,5) [1.008256] (6,6) [1.000655] (6,7) [1.004778] (6,8) [1.011365] (6,9) [1.012348] (6,10) [1.009210] (6,11) [1.010517] (6,12) [1.009383] (6,13) [1.008763] (6,14) [1.009113] (6,15) [1.012318] (6,16) [1.014063] (6,17) [1.009142] (6,18) [1.010196] (6,19) [1.007226] (6,20) [1.008919] 

(7,0) [1.006410] (7,1) [1.007096] (7,2) [1.012439] (7,3) [1.010103] (7,4) [1.009825] (7,5) [1.008970] (7,6) [1.001539] (7,7) [0.997641] (7,8) [1.010288] (7,9) [1.011264] (7,10) [1.008961] (7,11) [1.009354] (7,12) [1.009065] (7,13) [1.004591] (7,14) [1.010221] (7,15) [1.011112] (7,16) [1.014402] (7,17) [1.006413] (7,18) [1.010839] (7,19) [1.004666] (7,20) [1.002661] 

(8,0) [1.014035] (8,1) [1.011552] (8,2) [1.009015] (8,3) [1.006562] (8,4) [1.012678] (8,5) [1.007821] (8,6) [1.012583] (8,7) [1.012602] (8,8) [0.998037] (8,9) [1.011758] (8,10) [1.012278] (8,11) [1.010540] (8,12) [1.011573] (8,13) [1.012080] (8,14) [1.013852] (8,15) [1.011936] (8,16) [1.006914] (8,17) [1.010865] (8,18) [1.012645] (8,19) [1.012524] (8,20) [1.011191] 

(9,0) [1.007752] (9,1) [1.014632] (9,2) [1.015961] (9,3) [1.015311] (9,4) [1.014909] (9,5) [1.006350] (9,6) [1.010823] (9,7) [1.008276] (9,8) [1.014192] (9,9) [0.995093] (9,10) [1.014685] (9,11) [1.014463] (9,12) [1.014594] (9,13) [1.015198] (9,14) [1.015230] (9,15) [1.003974] (9,16) [1.014625] (9,17) [1.015570] (9,18) [1.015352] (9,19) [1.009127] (9,20) [1.014403] 

(10,0) [1.014198] (10,1) [1.007100] (10,2) [1.004849] (10,3) [1.008097] (10,4) [1.005460] (10,5) [1.012434] (10,6) [1.010296] (10,7) [1.011682] (10,8) [1.011758] (10,9) [1.013981] (10,10) [1.004590] (10,11) [1.012023] (10,12) [1.005802] (10,13) [1.009632] (10,14) [1.008600] (10,15) [1.013939] (10,16) [1.010857] (10,17) [1.008871] (10,18) [1.007212] (10,19) [1.014570] (10,20) [1.010268] 

(11,0) [1.006229] (11,1) [1.009443] (11,2) [1.009961] (11,3) [1.009250] (11,4) [1.011566] (11,5) [1.011293] (11,6) [1.009905] (11,7) [1.010210] (11,8) [1.009121] (11,9) [1.009897] (11,10) [1.010400] (11,11) [0.992653] (11,12) [1.010294] (11,13) [1.006829] (11,14) [1.008095] (11,15) [1.009897] (11,16) [1.006735] (11,17) [1.007086] (11,18) [1.009091] (11,19) [1.004643] (11,20) [1.005745] 

(12,0) [1.013738] (12,1) [1.008132] (12,2) [1.006412] (12,3) [1.007600] (12,4) [1.006772] (12,5) [1.012816] (12,6) [1.011593] (12,7) [1.012283] (12,8) [1.009524] (12,9) [1.013627] (12,10) [1.007046] (12,11) [1.011058] (12,12) [1.005730] (12,13) [1.009587] (12,14) [1.009431] (12,15) [1.013295] (12,16) [1.009653] (12,17) [1.009849] (12,18) [1.007567] (12,19) [1.013608] (12,20) [1.009994] 

(13,0) [1.005511] (13,1) [1.012267] (13,2) [1.011959] (13,3) [1.010084] (13,4) [1.012353] (13,5) [1.013004] (13,6) [1.012507] (13,7) [1.012326] (13,8) [1.007847] (13,9) [1.010735] (13,10) [1.011777] (13,11) [1.010630] (13,12) [1.011531] (13,13) [1.005652] (13,14) [1.012565] (13,15) [1.009564] (13,16) [1.012436] (13,17) [1.011442] (13,18) [1.012182] (13,19) [1.008814] (13,20) [1.009776] 

(14,0) [1.013442] (14,1) [1.010903] (14,2) [1.010535] (14,3) [1.013074] (14,4) [1.008425] (14,5) [1.012736] (14,6) [1.013069] (14,7) [1.013807] (14,8) [1.014065] (14,9) [1.012742] (14,10) [1.010078] (14,11) [1.011020] (14,12) [1.010770] (14,13) [1.012347] (14,14) [1.004922] (14,15) [1.012569] (14,16) [1.007919] (14,17) [1.012264] (14,18) [1.008432] (14,19) [1.013278] (14,20) [1.012542] 

(15,0) [1.011980] (15,1) [1.016948] (15,2) [1.016895] (15,3) [1.015632] (15,4) [1.016808] (15,5) [1.015826] (15,6) [1.015584] (15,7) [1.013745] (15,8) [1.012733] (15,9) [0.992399] (15,10) [1.017028] (15,11) [1.014698] (15,12) [1.016955] (15,13) [1.016401] (15,14) [1.016749] (15,15) [0.992130] (15,16) [1.017366] (15,17) [1.016042] (15,18) [1.016882] (15,19) [1.012227] (15,20) [1.015770] 

(16,0) [1.014228] (16,1) [1.012952] (16,2) [1.012536] (16,3) [1.015469] (16,4) [1.009875] (16,5) [1.013208] (16,6) [1.014727] (16,7) [1.015277] (16,8) [1.015590] (16,9) [1.012200] (16,10) [1.012216] (16,11) [1.014428] (16,12) [1.012992] (16,13) [1.014429] (16,14) [1.009003] (16,15) [1.011989] (16,16) [1.009821] (16,17) [1.014018] (16,18) [1.010666] (16,19) [1.013964] (16,20) [1.015037] 

(17,0) [1.007324] (17,1) [1.006513] (17,2) [1.007860] (17,3) [1.007301] (17,4) [1.009460] (17,5) [1.005462] (17,6) [1.003583] (17,7) [1.003227] (17,8) [1.009474] (17,9) [1.011404] (17,10) [1.007749] (17,11) [1.006867] (17,12) [1.007822] (17,13) [1.008207] (17,14) [1.009225] (17,15) [1.012208] (17,16) [1.006124] (17,17) [0.999405] (17,18) [1.008999] (17,19) [1.008169] (17,20) [1.004331] 

(18,0) [1.012891] (18,1) [1.005584] (18,2) [1.006221] (18,3) [1.010774] (18,4) [1.004566] (18,5) [1.007950] (18,6) [1.008646] (18,7) [1.011063] (18,8) [1.013851] (18,9) [1.013883] (18,10) [1.005373] (18,11) [1.010755] (18,12) [1.005779] (18,13) [1.011511] (18,14) [1.005001] (18,15) [1.014355] (18,16) [1.006888] (18,17) [1.009512] (18,18) [1.000923] (18,19) [1.012783] (18,20) [1.010450] 

(19,0) [1.005217] (19,1) [1.012212] (19,2) [1.011625] (19,3) [1.009905] (19,4) [1.013467] (19,5) [1.010881] (19,6) [1.011392] (19,7) [1.011418] (19,8) [1.007130] (19,9) [1.007540] (19,10) [1.012028] (19,11) [1.007083] (19,12) [1.011862] (19,13) [1.006924] (19,14) [1.012863] (19,15) [1.008012] (19,16) [1.011713] (19,17) [1.010130] (19,18) [1.012853] (19,19) [1.004173] (19,20) [1.009048] 

(20,0) [1.010634] (20,1) [1.009021] (20,2) [1.008849] (20,3) [1.007830] (20,4) [1.010001] (20,5) [1.011954] (20,6) [1.010703] (20,7) [1.011142] (20,8) [1.010304] (20,9) [1.011983] (20,10) [1.009120] (20,11) [1.005477] (20,12) [1.009088] (20,13) [1.009628] (20,14) [1.008801] (20,15) [1.011332] (20,16) [1.009406] (20,17) [1.008406] (20,18) [1.008950] (20,19) [1.011361] (20,20) [1.007979] 

};
\draw [draw=none,fill=blue!25] (3.5,-0.5) rectangle (4.5,20.500000);
\draw [draw=none,fill=blue!25] (12.5,-0.5) rectangle (13.5,20.500000);
\draw [draw=none,fill=blue!25] (15.5,-0.5) rectangle (16.5,20.500000);
\addplot[
    draw=none,
    fill=white,
    fill opacity=0.75
    ] coordinates {
    (3.000000,9.300000) (17.000000,10.700000) (17.000000,18.750000) (3.000000,1.250000)
};
\addplot[name path=A] coordinates {(3.000000,9.300000) (17.000000,10.700000)};
\addplot[name path=B] coordinates {(3.000000,1.250000) (17.000000,18.750000)};
\addplot[draw=none,fill=black] coordinates {(9.500000,9.500000) (10.500000,9.500000) (10.500000,10.500000) (9.500000,10.500000)};
\addplot[draw=blue] coordinates {(3.000000,3.000000) (17.000000,17.000000)};
\addplot[draw=blue] coordinates {(3.000000,7.666667) (17.000000,12.333333)};
\addplot[draw=blue] coordinates {(3.000000,6.500000) (17.000000,13.500000)};
\addplot[draw=none,fill=black] coordinates {(9.500000,9.500000) (10.500000,9.500000) (10.500000,10.500000) (9.500000,10.500000)};
\addplot[draw=none,fill=blue] coordinates {(3.500000,3.500000) (4.500000,3.500000) (4.500000,4.500000) (3.500000,4.500000)};
\addplot[draw=none,fill=blue] coordinates {(12.500000,10.500000) (13.500000,10.500000) (13.500000,11.500000) (12.500000,11.500000)};
\addplot[draw=none,fill=blue] coordinates {(15.500000,12.500000) (16.500000,12.500000) (16.500000,13.500000) (15.500000,13.500000)};

\end{axis}
\end{tikzpicture}

%% file: figs/method_uniqueness.tex
\begin{tikzpicture}
  \begin{axis}[
      height=0.5\columnwidth,
      width=\columnwidth,
      xlabel=Reference Image Number,
      ylabel=Reference Match Score,
      xtick pos=right,
      ticks=none,
      axis lines=left,
      y dir=reverse,
      xmin=0, xmax=16, ymin=0, ymax=5
    ]

    \draw[draw=none,fill=blue!25] (7.5,0.05) rectangle (12.5,5);

    \addplot [draw=none,mark=*] coordinates {
      (1,2.5)
      (2,2)
      (3,1.8)
      (4,2.25)
      (5,1.6)
      (6,2.4)
      (7,2.2)
      (8,3.6)
      (9,3.5)
      (10,4)
      (11,3.9)
      (12,3.7)
      (13,2.4)
      (14,2.8)
      (15,2.9)
      (16,2.1)
    };

    \addplot [dashed] coordinates { (0,4) (16,4) };
    \addplot [dashed] coordinates { (0,2.9) (16,2.9) };

    \addplot [mark=|] coordinates { (7.5,0.5) (12.5,0.5) };

    \node[] at (axis cs:0,2.9) [anchor=north west] {\small $\mu_2$};
    \node[] at (axis cs:0,4) [anchor=north west] {\small $\mu_1$};

    \node[] at (axis cs:10,0.5) [anchor=north] {\small $R_\mathrm{window}$};
  \end{axis}
\end{tikzpicture}

%% file: source/expdesign.tex
\section{Experimental Evaluation Design}
\label{sec:expdesign}

The OpenSeqSLAM2.0 toolbox was used to investigate and evaluate the influence of a number of parameters and variations in the SeqSLAM process. Sweeps of selected parameters were performed, for multiple datasets, under various configurations. The tests were selected to provide insights into the underlying behaviour of SeqSLAM, where much previous work has relied on ad hoc handcrafting and heuristics. The toolbox provides a batch operation mode out of the box, allowing users to easily examine the effects of specific parameters and configurations on performance with their own datasets.

\subsection{Datasets}

The parameter sweeps are performed on two benchmark datasets widely used in the literature \cite{sunderhauf2013we,sunderhauf2015performance,arroyo2015towards,siam2017fast,neubert2015local,vysotska2016lazy,chen2014convolutional,cummins2008fab}: Nordland and Eynsham.

\subsubsection{Nordland}

The dataset, first introduced in \cite{sunderhauf2013we}, comprises of footage of a 728km train journey captured across four different seasons. The dataset does not exhibit any viewpoint variation between traverses but presents severe changes in appearance due to seasonal variations. The experiments performed below used the two traverses with the most difference in conditions, that is, winter versus summer. As in \cite{sunderhauf2013we}, the traverses were subsampled such that frames were at 1fps and proposed matches were accepted if they were within 1 frame of the ground truth.

\subsubsection{Eynsham}

The dataset is a large 70km road-based dataset used by \cite{cummins2008fab} in the FAB-MAP studies. The dataset consists of two 35km traverses with panoramic images captured at 7m intervals by a Ladybug 2 camera. In the experiments below, the second traverse was matched to the first traverse. In accordance with the original FAB-MAP study (\cite{cummins2008fab}), the 40m GPS-derived ground truth provided with the Eynsham dataset was used for evaluating performance.

\subsection{Parameters}

Parameter sweeps were orchestrated to explore four areas of the SeqSLAM algorithm. The sweeps explored the enhancement of the difference matrix, length and method of sequence search through the matrix, and method for selecting from match proposals.

\subsubsection{Size of column normalisation window}

The sweep varied the width of the window $R_\mathrm{norm}$ used when enhancing the local contrast of the difference matrix (see Section \ref{subsec:normwidth} for parameter details). By varying the width, the sweep explored how limiting the window of normalisation affects maximum performance. The window width was varied from 2 to the length of traverse, with the basic trajectory search used (see Section \ref{subsec:search}), and match selection optimised for the best F1 score.

\subsubsection{Quantity of information used in searching}

Next, the length of sequences used in the search process (see $d_s$ in Section \ref{subsec:search}) was swept. By increasing the number of images in the search, the sweep investigated whether the intuition held for SeqSLAM that more information automatically wields better results. The length was varied from 2 to 40 frames, with the basic trajectory search used, and match selection optimised for the best F1 score.

\subsubsection{Method of searching for match candidates}

The relative performance of the three search methods described in Section \ref{subsec:search} was explored by sweeping the match selection threshold for each method. By focusing on the performance curve, and maximum F1 score, the sweep probed whether a best performing search method could be established. The sweep was performed across the range of valid basic match selection thresholds, for each of the search methods with a search length of 10 frames.

\subsubsection{Method of match selection from candidates}

Lastly, the effectiveness of each of the two search methods described in Section \ref{subsec:matches} was evaluated by sweeping their corresponding selection parameters. The shapes of the performance curves, and the maximum F1 score, were used to glean insights into which method performs the best and is most resistant to threshold selection error. The sweep was performed across the range of valid method thresholds, with the basic trajectory search used for each test.

%% file: source/results.tex
\section{Experimental Results and Discussion}
\label{sec:results}

The following section outlines the results for each of the four areas where parameter sweeps were performed. Plots for each area are presented, with comments on their defining features and identified trends discussed. For a number of the results plots (Figures \ref{fig:res_norm_width}, \ref{fig:res_search_nord}, \ref{fig:res_search_eyn}, \ref{fig:res_match_nord} and \ref{fig:res_match_eyns}), the valid value ranges do not match. Consequently, the plots are aligned by plotting the dependent axis as a percent of the tested value range.

\subsection{Normalisation width}

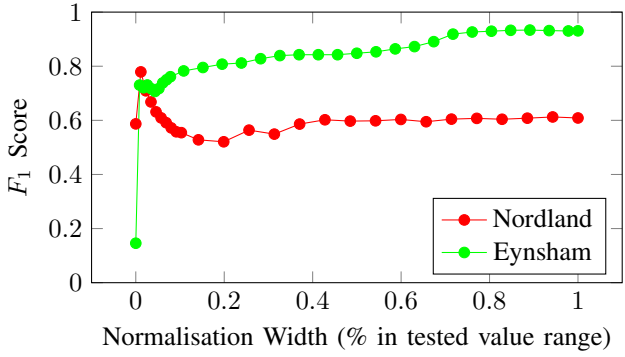
\begin{figure}[t]
  \centering
  \input{./figs/results_0_norm_width.tex}
  \caption{Effects of varying column normalisation width $R_\mathrm{norm}$ for the Nordland and Eynsham datasets.}
  \label{fig:res_norm_width}
\end{figure}

Normalisation width was varied between a minimum value of 2, and the maximum value equal to the number of reference images $n$ in the dataset. The results are shown in Figure \ref{fig:res_norm_width}, with the $F_1$ scores plotted for each of the datasets.

As can be seen, both datasets present an initial peak before dropping off and gradually increasing as more of the dataset is included in the normalisation. The initial peak for both datasets falls at around $2\%$ of the dataset length, which could be useful in situations where maximising performance for minimum compute cost is desired. Despite both peaking initially, the two dataset results differ in that Nordland never returns to its original peak, whereas Eynsham quickly surpasses it and continues to grow.

Reaching near-peak performance quickly, at a very small normalization window, has implications for designing computationally efficient and low latency systems. A small normalization window requires a shorter temporal history, which is especially relevant when performing initial global localisation at startup or after a kidnap event. The short window also is useful because it increases the applicability of the algorithm to highly fragmented environments where using a long normalization window over multiple different sections of the environment is undesirable.

\begin{figure}[t]
  \centering
  \input{./figs/results_1_seq_length.tex}
  \caption{Effects of varying sequence length $d_s$ for the Nordland and Eynsham datasets.}
  \label{fig:res_seq_length}
\end{figure}
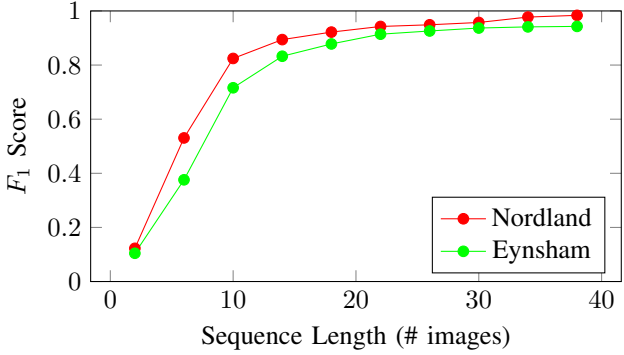

\subsection{Sequence Length}

Next, the results of varying sequence length between 2 and 40 frames for both datasets is presented. The results are shown in Figure \ref{fig:res_seq_length}, with both datasets presenting similar results. In both cases, performance improves as sequence length increases with diminishing performance returns the greater the sequence length.

The result of varying sequence length matches what intuition would suggest; the more information included in the search the greater the performance. It also highlights that the performance gains diminish as the amount of information provided to the search increases. For both datasets, a point of compromise (maximising performance while minimising information) exists for a sequence length of between 10 and 20.

\subsection{Search Method}

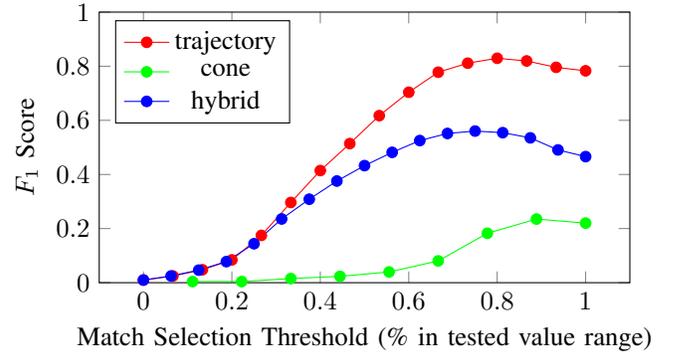
\begin{figure}[t]
  \centering
  \input{./figs/results_2_search_method_nord.tex}
  \caption{Characteristic performance curves for each search method with the Nordland dataset.}
  \label{fig:res_search_nord}
\end{figure}

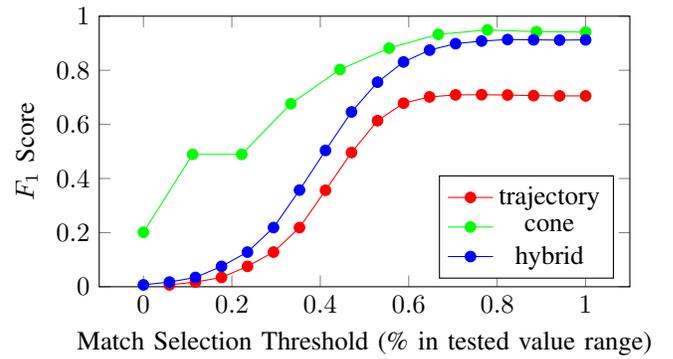
\begin{figure}[t]
  \centering
  \input{./figs/results_2_search_method_eyns.tex}
  \caption{Characteristic performance curves for each search method with the Eynsham dataset.}
  \label{fig:res_search_eyn}
\end{figure}

Thirdly, results from varying thresholds for each of the three search methods are presented for both datasets. Figure \ref{fig:res_search_nord} shows the results for the Nordland dataset, with the trajectory-based search performing best and cone-based search performing worst. Contrastingly, the results for the Eynsham dataset in Figure \ref{fig:res_search_eyn} show cone-based search as the best performer and trajectory-based as the weakest. The inversion in method performance can be explained by looking into the difference matrices for each dataset.

The Nordland matrix has no strong diagonal --- its global best matches are not along the ground truth diagonal due to the extreme condition variance --- until the matrix has its local contrast enhanced. Therefore the cone method will struggle to find correct matches because the global matches do not reflect the ground truth. In contrast, the global best matches for the Eynsham matrix do closely follow the ground truth due to the lack of condition variance. In understanding the varying performance of the trajectory-based search method, the explanation is more straightforward. The Nordland dataset has a straight diagonal ground truth, which fits linear trajectories, whereas the Eynsham dataset has a jagged diagonal ground truth to which linear trajectories cannot easily be fit.

Interestingly, all performance curves in the Nordland dataset decline after a peak whereas the Eynsham curves settle at the peak. The results suggest that the ability of a match selection to be too greedy (i.e. sacrifice precision) is a function of the dataset, rather than the search method. Therefore, it is concluded the most appropriate sequence matching method (from those presented in the literature) is highly dependent on the characteristics of the environment, and can be chosen according to the above observations.

\subsection{Match Selection Method}

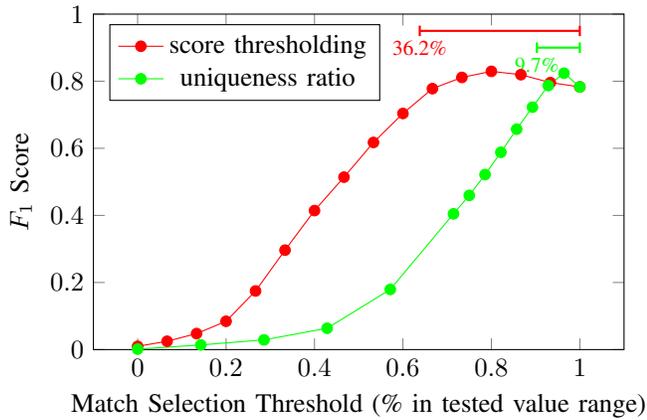
\begin{figure}[t]
  \centering
  \input{./figs/results_3_match_method_nord.tex}
  \caption{Characteristic performance curves for matching methods with the Nordland dataset. Performance bars denote $\%$ of value range for which performance $> 90\%$ of peak.}
  \label{fig:res_match_nord}
\end{figure}

Lastly, results from varying the match selection method are presented. For both the Nordland and Eynsham datasets, in Figures \ref{fig:res_match_nord} and \ref{fig:res_match_eyns} respectively, the maximum performance for the methods are similar. Where the methods differ is in the \textit{sensitivity of the method to threshold section}. The performance bars in the figures emphasise the difference in sensitivity, with the score thresholding producing a $4 - 6 \times$ wider stable performance parameter range.

The score thresholding method is much more tolerant to threshold selection, whereas the windowed uniqueness method has a small peak around the maximum value with performance rapidly deteriorating as $\mu$ selection moves away from the peak. Consequently, the score thresholding method appears equally effective from a performance point of view, and more stable under parameter tweaking.

\begin{figure}[h!]
  \centering
  \input{./figs/results_3_match_method_eyns.tex}
  \caption{Characteristic performance curves for matching methods with the Eynsham dataset. Performance bars denote $\%$ of value range for which performance $> 90\%$ of peak.}
  \label{fig:res_match_eyns}
\end{figure}

%% file: figs/results_0_norm_width.tex
    \begin{tikzpicture}
    \begin{axis}[
    width=\columnwidth,
    height=0.6\columnwidth,
xlabel=Normalisation Width (\% in tested value range),
legend pos=south east,
ymin=0, ymax=1,
ylabel=$F_1$ Score
]
\addplot[red, mark=*] coordinates { (0.000000, 0.586957) (0.011445, 0.779032) (0.022890, 0.709076) (0.034335, 0.668013) (0.045780, 0.631368) (0.057225, 0.609113) (0.068670, 0.591385) (0.080114, 0.572091) (0.091559, 0.557685) (0.103004, 0.554424) (0.141631, 0.527975) (0.198856, 0.520701) (0.256080, 0.563815) (0.313305, 0.549088) (0.370529, 0.586193) (0.427754, 0.602028) (0.484979, 0.597199) (0.542203, 0.598269) (0.599428, 0.603292) (0.656652, 0.594817) (0.713877, 0.604488) (0.771102, 0.607293) (0.828326, 0.604032) (0.885551, 0.607994) (0.942775, 0.612479) (1.000000, 0.608696) };
\addplot[green, mark=*] coordinates { (0.000000, 0.145854) (0.008699, 0.730240) (0.017399, 0.720446) (0.026098, 0.730598) (0.034798, 0.717044) (0.043497, 0.706752) (0.052197, 0.717508) (0.060896, 0.738279) (0.069595, 0.750030) (0.078295, 0.761206) (0.108308, 0.782619) (0.151805, 0.795260) (0.195302, 0.807613) (0.238799, 0.811868) (0.282297, 0.827736) (0.325794, 0.839257) (0.369291, 0.842517) (0.412788, 0.842517) (0.456285, 0.842517) (0.499783, 0.847951) (0.543280, 0.853245) (0.586777, 0.864080) (0.630274, 0.872501) (0.673771, 0.890773) (0.717268, 0.918596) (0.760766, 0.926421) (0.804263, 0.929464) (0.847760, 0.932725) (0.891257, 0.933594) (0.934754, 0.931638) (0.978251, 0.930334) (1.000000, 0.930768) };
\legend {Nordland, Eynsham}
    \end{axis}
    \end{tikzpicture}
    

%% file: figs/results_1_seq_length.tex
    \begin{tikzpicture}
    \begin{axis}[
    width=\columnwidth,
    height=0.6\columnwidth,
xlabel=Sequence Length (\# images),
legend pos=south east,
ymin=0, ymax=1,
ylabel=$F_1$ Score
]
\addplot[red, mark=*] coordinates { (2.000000, 0.122705) (6.000000, 0.530684) (10.000000, 0.824421) (14.000000, 0.894091) (18.000000, 0.921805) (22.000000, 0.942605) (26.000000, 0.948830) (30.000000, 0.957408) (34.000000, 0.977369) (38.000000, 0.983711) };
\addplot[green, mark=*] coordinates { (2.000000, 0.104496) (6.000000, 0.375883) (10.000000, 0.716020) (14.000000, 0.832598) (18.000000, 0.878020) (22.000000, 0.914080) (26.000000, 0.925939) (30.000000, 0.936834) (34.000000, 0.941096) (38.000000, 0.942893) };
\legend {Nordland, Eynsham}
    \end{axis}
    \end{tikzpicture}
    

%% file: figs/results_2_search_method_nord.tex
    \begin{tikzpicture}
    \begin{axis}[
    width=\columnwidth,
    height=0.6\columnwidth,
xlabel=Match Selection Threshold (\% in tested value range),
legend pos=north west,
ymin=0, ymax=1,
ylabel=$F_1$ Score
]
\addplot[red, mark=*] coordinates { (0.000000, 0.009736) (0.066667, 0.024845) (0.133333, 0.047749) (0.200000, 0.084337) (0.266667, 0.174745) (0.333333, 0.296429) (0.400000, 0.414404) (0.466667, 0.514019) (0.533333, 0.617391) (0.600000, 0.703804) (0.666667, 0.777872) (0.733333, 0.811268) (0.800000, 0.829400) (0.866667, 0.819429) (0.933333, 0.795867) (1.000000, 0.783035) };
\addplot[green, mark=*] coordinates { (0.000000, nan) (0.111111, 0.004184) (0.222222, 0.004184) (0.333333, 0.015257) (0.444444, 0.023481) (0.555556, 0.039726) (0.666667, 0.080483) (0.777778, 0.182638) (0.888889, 0.235152) (1.000000, 0.220119) };
\addplot[blue, mark=*] coordinates { (0.000000, 0.009736) (0.062500, 0.024845) (0.125000, 0.046416) (0.187500, 0.077905) (0.250000, 0.143969) (0.312500, 0.235512) (0.375000, 0.308511) (0.437500, 0.375709) (0.500000, 0.432640) (0.562500, 0.481698) (0.625000, 0.525232) (0.687500, 0.551482) (0.750000, 0.560353) (0.812500, 0.554383) (0.875000, 0.535247) (0.937500, 0.490638) (1.000000, 0.465921) };
\legend {trajectory, cone, hybrid}
    \end{axis}
    \end{tikzpicture}
    

%% file: figs/results_2_search_method_eyns.tex
    \begin{tikzpicture}
    \begin{axis}[
    width=\columnwidth,
    height=0.6\columnwidth,
xlabel=Match Selection Threshold (\% in tested value range),
legend pos=south east,
ymin=0, ymax=1,
ylabel=$F_1$ Score
]
\addplot[red, mark=*] coordinates { (0.000000, nan) (0.058824, 0.006969) (0.117647, 0.017331) (0.176471, 0.034364) (0.235294, 0.075289) (0.294118, 0.128043) (0.352941, 0.218720) (0.411765, 0.356503) (0.470588, 0.496034) (0.529412, 0.613810) (0.588235, 0.678313) (0.647059, 0.701617) (0.705882, 0.709101) (0.764706, 0.709599) (0.823529, 0.708534) (0.882353, 0.706279) (0.941176, 0.705358) (1.000000, 0.705281) };
\addplot[green, mark=*] coordinates { (0.000000, 0.201258) (0.111111, 0.489270) (0.222222, 0.489270) (0.333333, 0.676500) (0.444444, 0.802824) (0.555556, 0.881851) (0.666667, 0.932657) (0.777778, 0.949043) (0.888889, 0.942521) (1.000000, 0.941317) };
\addplot[blue, mark=*] coordinates { (0.000000, 0.006969) (0.058824, 0.017331) (0.117647, 0.034364) (0.176471, 0.075289) (0.235294, 0.128043) (0.294118, 0.218762) (0.352941, 0.357143) (0.411765, 0.503757) (0.470588, 0.645932) (0.529412, 0.756101) (0.588235, 0.830734) (0.647059, 0.874491) (0.705882, 0.898266) (0.764706, 0.908047) (0.823529, 0.913575) (0.882353, 0.912511) (0.941176, 0.911370) (1.000000, 0.912334) };
\legend {trajectory, cone, hybrid}
    \end{axis}
    \end{tikzpicture}
    

%% file: figs/results_3_match_method_nord.tex
    \begin{tikzpicture}
    \begin{axis}[
    width=\columnwidth,
    height=0.7\columnwidth,
xlabel=Match Selection Threshold (\% in tested value range),
legend pos=north west,
ymin=0, ymax=1,
ylabel=$F_1$ Score
]
\addplot[red, mark=*] coordinates { (0.000000, 0.009736) (0.066667, 0.024845) (0.133333, 0.047749) (0.200000, 0.084337) (0.266667, 0.174745) (0.333333, 0.296429) (0.400000, 0.414404) (0.466667, 0.514019) (0.533333, 0.617391) (0.600000, 0.703804) (0.666667, 0.777872) (0.733333, 0.811268) (0.800000, 0.829400) (0.866667, 0.819429) (0.933333, 0.795867) (1.000000, 0.783035) };
\addplot[green, mark=*] coordinates { (1.000000, 0.783035) (0.964286, 0.823811) (0.928571, 0.787037) (0.892857, 0.722569) (0.857143, 0.656969) (0.821429, 0.588061) (0.785714, 0.521694) (0.750000, 0.459634) (0.714286, 0.404682) (0.571429, 0.179389) (0.428571, 0.063599) (0.285714, 0.028926) (0.142857, 0.013879) (0.000000, 0.001397) };
\legend {score thresholding, uniqueness ratio}\addplot[red,mark=|,thick] coordinates {(0.638,0.95) (1.0,0.95) };
\addplot[green,mark=|,thick] coordinates {(0.903,0.9) (1.0,0.9) };
\node [text=red] at (axis cs:0.638,0.95) [anchor=north] {\footnotesize 36.2\%};
\node [text=green] at (axis cs:0.903,0.9) [anchor=north] {\footnotesize 9.7\%};

    \end{axis}
    \end{tikzpicture}
    

%% file: figs/results_3_match_method_eyns.tex
    \begin{tikzpicture}
    \begin{axis}[
    width=\columnwidth,
    height=0.7\columnwidth,
xlabel=Match Selection Threshold (\% in tested value range),
legend pos=north west,
ymin=0, ymax=1,
ylabel=$F_1$ Score
]
\addplot[red, mark=*] coordinates { (0.000000, nan) (0.058824, 0.006969) (0.117647, 0.017331) (0.176471, 0.034364) (0.235294, 0.075289) (0.294118, 0.128043) (0.352941, 0.218720) (0.411765, 0.356503) (0.470588, 0.496277) (0.529412, 0.613810) (0.588235, 0.678313) (0.647059, 0.701617) (0.705882, 0.709101) (0.764706, 0.709599) (0.823529, 0.708534) (0.882353, 0.706279) (0.941176, 0.705358) (1.000000, 0.705281) };
\addplot[green, mark=*] coordinates { (1.000000, 0.705281) (0.964286, 0.716966) (0.928571, 0.654702) (0.892857, 0.564919) (0.857143, 0.477235) (0.821429, 0.389729) (0.785714, 0.311128) (0.750000, 0.251433) (0.714286, 0.198070) (0.571429, 0.083752) (0.428571, 0.033520) (0.285714, 0.015181) (0.142857, 0.007836) (0.000000, 0.001747) };
\legend {score thresholding, uniqueness ratio}\addplot[red,mark=|,thick] coordinates {(0.5521,0.95) (1.0,0.95) };
\addplot[green,mark=|,thick] coordinates {(0.9232,0.9) (1.0,0.9) };
\node [text=red] at (axis cs:0.5521,0.95) [anchor=north west] {\footnotesize 44.8\%};
\node [text=green] at (axis cs:0.9232,0.9) [anchor=north] {\footnotesize 7.7\%};

    \end{axis}
    \end{tikzpicture}
    

%% file: source/conclusion.tex
\section{Discussion and Conclusions}
\label{sec:conclusion}

In summary, the paper has introduced the OpenSeqSLAM2.0 toolbox as novel open source software for comprehensively characterising the performance relationship between the key SeqSLAM system components and parameters to maximise visual place recognition performance. The toolbox's batch parameter sweep mode was used to investigate the effects of varying column normalisation width, sequence length, sequence search method, and match candidate selection method. From the results, conducted on two established datasets, the following conclusions were observed:
\begin{itemize}
  \item A column normalisation width of $2\%$ of the dataset size presents a performance peak, before large width increases are required to further increase performance. The optimal performance of a short window has positive implications for initial global localisation, highly fragmented environments, and low latency systems.
  \item A sequence length of between 10 and 20 frames maximises performance while not unnecessarily increasing the algorithm's computation costs. Significantly, the optimal operating range manages to maximise performance while minimising computational cost.
  \item The hybrid sequence search method has the most stable performance, with trajectory-based and cone-based search performance dependent on the shape of the dataset ground truth and condition variance respectively. Results suggest the trajectory-based search methods will perform better in environments with high condition variance and consistent dataset velocity, whereas cone-based methods perform best for visually similar traverses with inconsistent dataset velocity.
  \item The score thresholding method is $4 - 6 \times$ less sensitive to threshold selection than the windowed uniqueness method, with similar peak performance. The result suggest score thresholding should be preferred by default.
\end{itemize}

Although deep learning-based techniques have come to the fore recently \cite{chen2014convolutional,milford2015sequence,chen2017only,neubert2015local,sunderhauf2015performance,bai2017cnn,lowry2016visual}; there is still a significant role for traditional visual place recognition techniques, especially in applications where compute is constrained, sufficient training data is not always available, or in combination with challenging environmental conditions such as in autonomous underground mining vehicles \cite{zeng2017enhancing}. Sequence-based techniques are also relevant even when deep-learnt image representations are involved, because they can be added on top of single image based retrieval to further improve performance. The open source toolbox enhances the ability of the research community to work with SeqSLAM by providing automatic threshold optimisation, dynamic parameter reconfiguration, batch operation, performance characterisation, and other assisting features. This paper for the first time comprehensively characterised many of the critical parameters of the SeqSLAM system, providing an evaluation of their utility and performance which builds on top of mostly ad hoc former implementations. By providing for the first time a comprehensive open source software package that enables a complete characterization of one of the most widely used and benchmarked-against techniques SeqSLAM, the aim is to further facilitate research in the growing topical area of visual place recognition.